\documentclass[runningheads]{llncs}

\usepackage{eccv}

\usepackage{eccvabbrv}

\usepackage{graphicx}
\usepackage{booktabs}
\newcommand\blfootnote[1]{%
  \begingroup
  \renewcommand\thefootnote{}\footnote{#1}%
  \addtocounter{footnote}{-1}%
  \endgroup
}
\usepackage[accsupp]{axessibility}  % Improves PDF readability for those with disabilities.

\usepackage[pagebackref,breaklinks,colorlinks,citecolor=eccvblue]{hyperref}
\usepackage{orcidlink}

\begin{document}

% ---------------------------------------------------------------
% TODO REVIEW: Replace with your title
\title{RoomTex: Texturing Compositional Indoor Scenes via Iterative Inpainting} 

% TODO REVIEW: If the paper title is too long for the running head, you can set
% an abbreviated paper title here. If not, comment out.
\titlerunning{RoomTex}

% TODO FINAL: Replace with your author list. 
% Include the authors' OCRID for the camera-ready version, if at all possible.
\author{Qi Wang\inst{1*} \and
Ruijie Lu\inst{2*} \and
Xudong Xu\inst{3} \and
Jingbo Wang\inst{3} \and
Michael Yu Wang\inst{1}, \\
Bo Dai\inst{3\dag} \and
Gang Zeng\inst{2} \and
Dan Xu\inst{1}
}

% TODO FINAL: Replace with an abbreviated list of authors.
\authorrunning{Q. Wang et al.}
% First names are abbreviated in the running head.
% If there are more than two authors, 'et al.' is used.

% TODO FINAL: Replace with your institution list.
\institute{The Hong Kong University of Science and Technology \and
Peking University \and Shanghai AI Laboratory
% Princeton University, Princeton NJ 08544, USA \and
% Springer Heidelberg, Tiergartenstr.~17, 69121 Heidelberg, Germany
% \email{lncs@springer.com}\\
% \url{http://www.springer.com/gp/computer-science/lncs} \and
% ABC Institute, Rupert-Karls-University Heidelberg, Heidelberg, Germany\\
% \email{\{abc,lncs\}@uni-heidelberg.de}
}

\maketitle
\blfootnote{
\noindent $*$ Equal contribution, work done during the internship at Shanghai AI Laboratory. \\
$\dag$ Corresponding author: \href{mailto:daibo@pjlab.org.cn}{\color{black}{daibo@pjlab.org.cn}}
}

% !TEX root = ../main.tex

\begin{abstract}
% Creating finely textured 3D scenes is a promising yet challenging research direction with tremendous potential for application in games, AR/VR, etc. Recent advances have been made in the realm of object geometry and texture generation.
%     3D indoor scene generation is a promising but challenging research direction with tremendous potential for application in games, AR/VR, etc. 
% Recent works have either tried to incrementally generate an indoor scene or generate a panoramic image to represent it,
% but suffer from 3D inconsistency, poor geometry, limited view angles, and so on.

The advancement of diffusion models has pushed the boundary of text-to-3D object generation.
While it is straightforward to composite objects into a scene with reasonable geometry,
it is nontrivial to texture such a scene perfectly due to style inconsistency and occlusions between objects. 
To tackle these problems, we propose a \emph{coarse-to-fine} 3D scene texturing framework, referred to as \textbf{RoomTex},
to generate high-fidelity and style-consistent textures for untextured compositional scene meshes.
% The key of \textbf{RoomTex} is a two-stage texturing process.
%
% Given an assembled room generated by object generators and a layout as input, \textbf{RoomTex} first leverages ControlNet to generate a panoramic image as the coarse reference to ensure the global texture consistency of the room in the coarse stage.
In the coarse stage, RoomTex first unwraps the scene mesh to a panoramic depth map and leverages ControlNet to generate a room panorama,
which is regarded as the coarse reference to ensure the global texture consistency.
In the fine stage,
based on the panoramic image and perspective depth maps, RoomTex will refine and texture every single object in the room iteratively along a series of selected camera views,
until this object is completely painted.
Moreover, we propose to maintain superior alignment between RGB and depth spaces via subtle edge detection methods.
Extensive experiments show our method is capable of generating high-quality and diverse room textures,
and more importantly,
supporting interactive fine-grained texture control and flexible scene editing
thanks to our inpainting-based framework and compositional mesh input.
% our method can support interactive fine-grained texture control and flexible scene editing. 
Our project page is available at \href{https://qwang666.github.io/RoomTex/}{https://qwang666.github.io/RoomTex/}.

\end{abstract}
% !TEX root = ../main.tex

\section{Introduction}
\label{sec:intro}

\begin{figure*}[t!]
	\centering
	\includegraphics[width=0.9\linewidth]{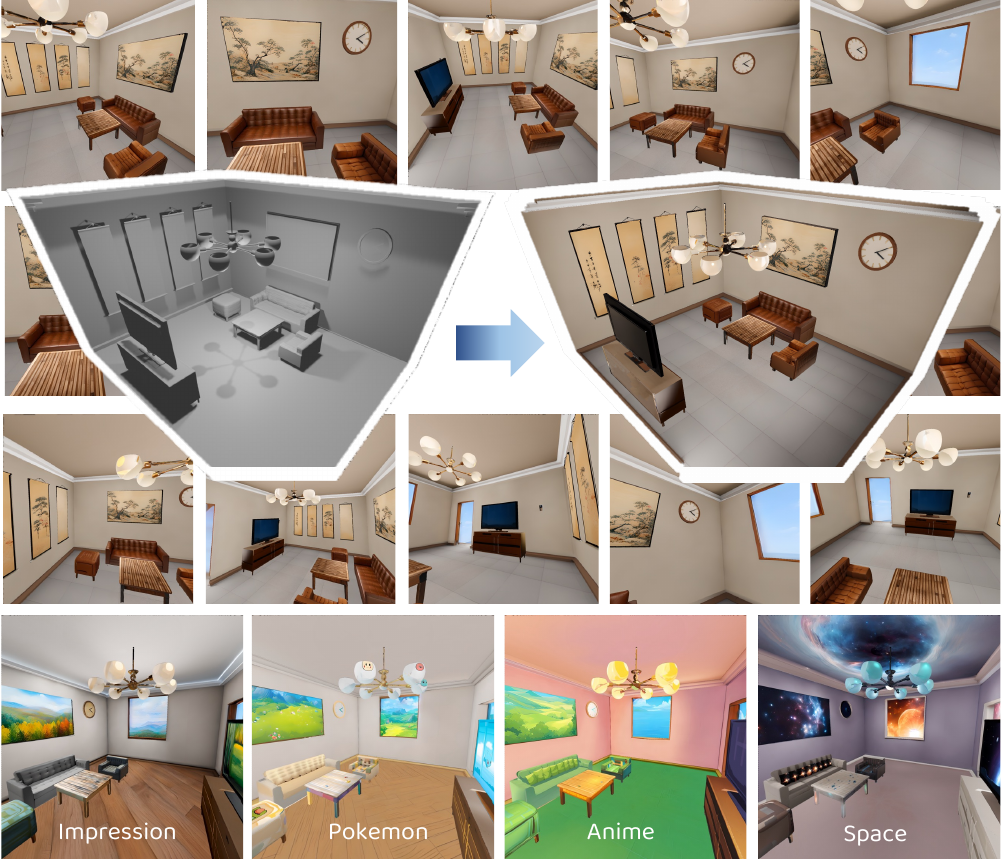}
	\caption{\small We propose \textbf{RoomTex} to synthesize high-quality and style-consistent textures for given scene meshes. Our method supports generating multiple styles.}
    \label{fig:teaser}
\end{figure*}

Generating high-quality textured 3D models, especially indoor scenes, is imperative for various industrial applications, ranging from gaming and filming to AR/VR.
Current delicate 3D indoor scenes, however, are mostly carefully designed by professional artists with expertise and thus is an expensive and time-consuming process.
Recently, significant progress has been made in the realm of 3D object generation~\cite{poole2022dreamfusion,chen2023fantasia3d,li2023sweetdreamer,qiu2023richdreamer,siddiqui2023meshgpt,gao2022get3d,gupta20233dgen,liu2023meshdiffusion}, especially in terms of geometry quality.
Despite satisfactory scene geometry, achieving captivating and style-consistent textures still demands painstaking efforts from artists equipped with specialized knowledge and aesthetic training.
%Yet, simply extending these methods to 3D indoor scene generation is insufficient to deal with the complicated occlusion problems while maintaining style consistency between all the components in the scene.
% Yet, it is nontrivial to extend these methods to 3D indoor scene generation, especially synthesizing compelling and style-consistent texture, owing to the complicated occlusion problems in the scene.
% Although an indoor scene can be viewed as the composition of various 3D objects,
% it is nontrivial to ensure compelling and globally style-consistent texture for all the individual objects in the scene,
Hence, automatic scene texturing, \ie, generating textures for untextured scene-level meshes, remains a valuable but challenging problem.

Existing texturing methods mostly focus on synthesizing textures for 3D objects, most of which are either limited to several specific training categories~\cite{oechsle2019texture,siddiqui2022texturify,bokhovkin2023mesh2tex} or relying on the corresponding UV maps~\cite{chen2022auv,yu2023texture}.
Thanks to powerful CLIP model~\cite{radford2021learning} connecting text and images,
some subsequent approaches~\cite{text2mesh,chen2022tango,yu2023texture,chen2023text2tex,cao2023texfusion,zeng2023paint3d} is capable of painting \emph{general} 3D objects by leveraging CLIP model or more advanced text-to-image diffusion models.
Conditioned on given text descriptions, they typically apply an iterative scheme to texture an object from different viewpoints.
Despite remarkable results on 3D objects, these methods cannot be naively extended to 3D indoor scenes owing to the complicated occlusion problem in the scene.
Although an indoor scene can be viewed as the composition of various textured objects,
it is nontrivial to ensure global style consistency of texture for all the individual objects inside.

In this work, we propose a novel \emph{coarse-to-fine} framework, dubbed \textbf{RoomTex}, to synthesize high-fidelity and style-consistent texture for a compositional 3D indoor scene under the guidance of text prompt that simultaneously enables flexible scene editing and fine-grained texture control. 
% Our key insight is to decompose the generation of 3D indoor scenes into geometry and texture generation.
Contrary to directly using perfect indoor scene meshes~\cite{fu20213d} designed by professional artists,
we opt to leverage off-the-shelf 3D object generative models along with a given 2D room layout to form a compositional untextured scene mesh with imperfect geometry, which alleviates the time-consuming model-making process and better aligns with the great development of 3D generative models.
Afterward, this indoor scene mesh is firstly unwrapped to obtain a global panoramic depth map. Based on the depth map and text prompt, RoomTex leverages ControlNet~\cite{zhang2023adding} to synthesize a panoramic image of the entire scene in the coarse stage.
It is noteworthy that such a room panorama is regarded as the coarse reference in the subsequent fine stage to maintain global style consistency.
To cope with the occlusion problem between objects and interior surfaces, we will remove all the objects inside to inpaint the occluded areas to acquire complete room interior surfaces, including walls, floor, and ceilings.

% that possess perfect and smooth geometry, we opt to apply an off-the-shelf object generator along with a given 2D layout to acquire an untextured compositional scene. Though imperfect, these compositional scenes can alleviate the time-consuming process of building a professional dataset. Afterwards, \textbf{RoomTex} connects scene geometry and texture via panoramic images. 
% % While geometry generation is achieved by applying an off-the-shelf object generator conditioned on a given 2D layout,
% % \textbf{RoomTex} connects scene geometry and texture via panoramic images.
% Specifically, to obtain high-quality scene texture,
% a panoramic image of the entire room is synthesized via ControlNet~\cite{zhang2023adding} under the guidance of a global depth map as the coarse reference,
% based on which we further refine and inpaint every single 3D object using depth rendered from multiple novel viewpoints.
% % It's noteworthy that our depth guidance can be obtained from a collection of object meshes, even unsmooth meshes generated by imperfect 3D generative models.
% To ensure compositional generation, we will remove all the objects afterward to inpaint the occluded area to acquire complete room interior surfaces, including walls, floor, and ceilings.

\begin{figure*}[t!]
	\centering
	\includegraphics[width=0.9\linewidth]{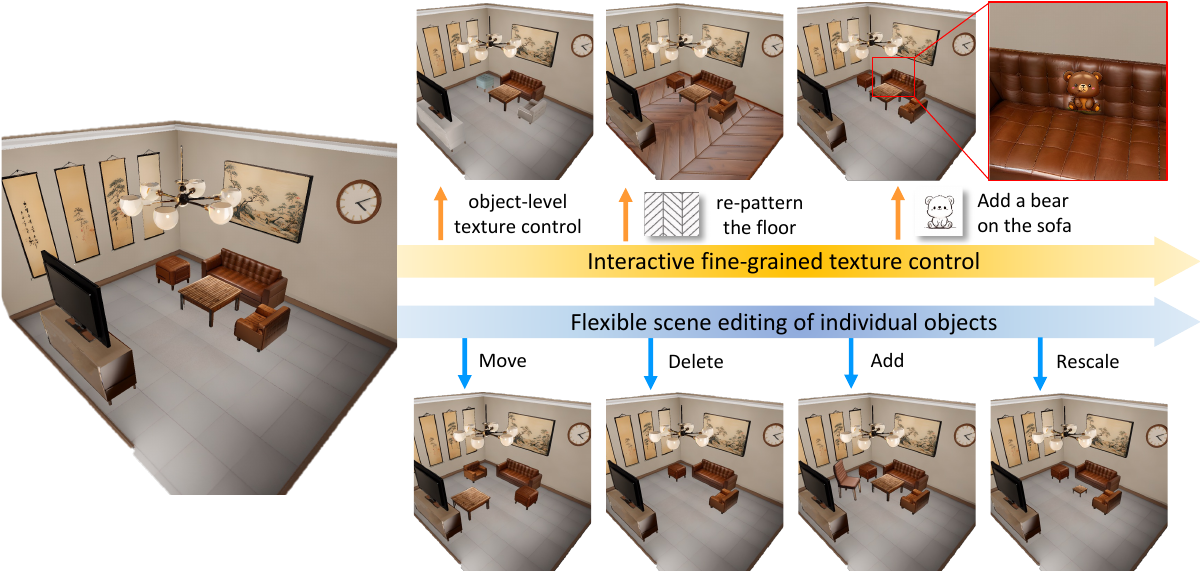}
	\caption{\small \textbf{RoomTex} simultaneously enables interactive fine-grained texture control and flexible scene editing of individual objects inside. }
    \label{fig:teaser-control}
\end{figure*}

In the fine stage, RoomTex will texture every single object in the room based on the panoramic image acquired in the coarse stage, leading to a complete 3D room that can be perceived from any viewpoint.
Specifically, for a particular 3D object to be painted, we first re-project the panorama from an appropriate view targeting this object to obtain a perspective image of the object.
However, this perspective image inevitably contains unacceptable distortions owing to the equirectangular projection and thus will be refined to a better and more detailed image via depth-guided ControlNet~\cite{zhang2023adding}.
Apart from this initial view, we additionally select a series of camera positions around this 3D object, along which this 3D object will be textured iteratively under the guidance of text prompt.
Regarding the refined perspective image as a starting point,
we warp this partially painted 3D object to other viewpoints under the depth guidance and inpaint the missing texture in each new viewpoint, which repeats until the 3D object is completely painted.
Unfortunately, the generated texture cannot perfectly align with the guiding depth map.
In particular, the texture in the foreground of an object will dilate to the background area, which typically occurs in the depth edge areas and might be imperceptible from the current view.
However, the dilated texture from the previous iteration leads to a messy area while the object is warped to a new viewpoint,
which will be exacerbated as the iterative inpainting goes on.
To mitigate this problem, we propose to detect these misalignment areas with Canny~\cite{canny1986computational} edges of RGB images and Laplacian edges of the corresponding depth maps.
Afterward, these areas will not taken into consideration during the iterative inpainting to avoid awful object textures.

Extensive experiments demonstrate that our method can synthesize high-fidelity and style-consistent texture for a compositional room mesh conditioned on the given text prompts as shown in Fig.~\ref{fig:teaser}.
% More importantly, all components in our generated scene are naturally compositional, and thus users can flexibly edit any independent 3D object in the scene.
More importantly, thanks to the powerful control capabilities of ControlNet~\cite{zhang2023adding} and our inpainting-based texturing framework,
RoomTex supports flexible scene editing of all individual objects and fine-grained texture control such as aligning the texture of a specific 3D object with provided sketches or text descriptions as illustrated in Fig.~\ref{fig:teaser-control}.
Our contributions can be summarized as follows:
\begin{itemize}
    \item We propose a novel coarse-to-fine texture generation framework that first generates a room panorama as a coarse reference and then paints each component in the scene to achieve global style consistency.
    \item Thanks to our subtly designed alignment between RGB and depth spaces, our method can take imperfect geometry from 3D object generative models as input and generate holistic and high-fidelity scene textures.
    % \item We disentangle all the components in the scene, providing users with a highly editable room where they can add, remove, replace, move, and rescale any furniture item inside.
    % \item Our method can support fine-grained controls such as scene editing, aligning textures with sketches and text well due to the explicit texturing framework, providing users with a highly flexible room.
    \item Users can not only flexibly edit the indoor scene where they can add, remove, replace, move, and rescale any furniture item, but also realize fine-grained texture control over any object with given sketches or text prompts.
\end{itemize}

% !TEX root = ../main.tex

\section{Related Work}
\label{sec:relwork}

\noindent\textbf{Diffusion-Based Text-to-Image Generation.}
In recent years, diffusion probabilistic models~\cite{ho2020denoising,sohl2015deep} have achieved unprecedented success in text-to-image generation.
By training on large-scale text-image paired datasets~\cite{sharma2018conceptual, schuhmann2021laion}, diffusion models manage to learn an implicit connection between semantic concepts and corresponding text embeddings, thus generating diverse and complex images of objects and scenes from given text prompts~\cite{nichol2022glide,radford2021learning,saharia2022photorealistic,balaji2022ediffi}.
Different from pixel-based diffusion approaches,
latent diffusion models~\cite{rombach2022high} (LDMs) apply the diffusion model on the latent space of pretrained autoencoders, significantly reducing the demands for massive computational resources.
Moreover, several fantastic works~\cite{zhang2023adding, mou2023t2i, voynov2023sketch} have explored utilizing additional conditions like sketches, Canny edges, depth maps, \etc, to control the image generation of large pretrained text-to-image models.
In this work, we leverage ControlNet~\cite{zhang2023adding} to synthesize indoor panorama conditioned on the corresponding indoor depth map and subsequent inpainting or editing procedures.

% \vspace{-0.15in}

\noindent\textbf{Text-Driven 3D Object Generation and Texturing.}
The great success of text-to-image synthesis empowers booming development in the domain of text-to-3D generation.
Based on powerful 2D text-to-image diffusion models, DreamFusion~\cite{poole2022dreamfusion} first proposes an effective Score Distillation Sampling (SDS) loss to guide the generation of 3D models.
Later, a vast body of following works leverages the SDS loss to synthesize various 3D objects with higher quality and better 3D consistency~\cite{wang2023score,lin2023magic3d,chen2023fantasia3d,wang2023prolificdreamer,shi2023mvdream,li2023sweetdreamer,liu2023unidream,qiu2023richdreamer}.
By training on massive 3D synthetic data, prior attempts~\cite{nichol2022point,jun2023shap} on direct generation of 3D point clouds or meshes are shown to significantly accelerate the generation process.
%Some approaches~\cite{nichol2022point,jun2023shap} also attempt to directly generate 3D point clouds or meshes from the given text prompts by training on massive 3D synthetic data.
It is noteworthy that our method can capitalize on these generated 3D objects for room composition as input, despite their imperfect geometry.
Given untextured 3D meshes and the conditioning text descriptions or reference images,
some approaches~\cite{richardson2023texture,chen2023text2tex,michel2022text2mesh,wang2023breathing,zeng2023paint3d,yeh2024texturedreamer,yu2023texture,cao2023texfusion,youwang2023paint,metzer2023latent} also exploit the text-to-image diffusion models for 3D object texturing by using an iterative painting scheme or relying on the corresponding UV map.
Unlike them, we aim to paint the entire room, including each independent 3D object inside, with high-quality and consistent textures.

% \vspace{-0.15in}

% \noindent\textbf{Room Layout Synthesis.}
% A reasonable room layout is the prerequisite for generating a realistic 3D room.
% Several methods~\cite{paschalidou2021atiss,tang2023diffuscene} are capable of generating suitable room layouts by learning from professional designs~\cite{fu20213d}.
% %
% However, this line of research works only considers relative positions of furniture while ignoring surroundings including walls, floors, and ceilings, which are indispensable parts of a complete 3D room.
% % Moreover, inconsistent texture styles between furniture may appear while rearranging these pre-designed furniture models.
% Our method relies on a pre-defined or generated room layout but targets generating a complete 3D room.

% \vspace{-0.15in}

\noindent\textbf{Indoor Scene Generation with Panorama.}
Recently, MVDiffusion~\cite{tang2023mvdiffusion} and Ctrl-Room~\cite{fang2023ctrl} subtly design their specific diffusion models to synthesize multi-view consistent images or 3D layouts of indoor scenes.
However, they are both constrained to generating a panoramic image to represent the whole 3D room and thus wandering around the room is far beyond the capability of these methods.
Although Ctrl-Room further combines the estimated depth map with the panoramic image to obtain a complete textured room,
the occlusion area cannot be covered with a single panorama and the potential panoramic distortion remains out of reach.
In contrast, our method aims to texture a compositional room mesh and leverages a panorama to ensure style consistency in the scene.

% \vspace{-0.15in}

\noindent\textbf{Text-Driven 3D Scene Indoor Generation and Texturing.}
Several prior works~\cite{cohen2023set, po2023compositional} explore generating a 3D indoor scene by using 3D bounding boxes as layouts and optimizing the entire scene with SDS loss~\cite{poole2022dreamfusion}.
Yet, the texure quality is still far from satisfactory since the generated scene often looks unrealistic and over-saturated.
Alternative approaches~\cite{hollein2023text2room, zhang2023text2nerf, fridman2023scenescape} adopt an incremental framework,
where they mainly leverage image warping to obtain renderings from new viewpoints and then inpaint the missing areas based on the estimated depth map.
However, inaccurate depth estimation leads to severe geometry distortion, significantly affecting the generation results.
% the depth estimated with monocular cues is always inaccurate and inconsistent.
% Therefore, the obtained 3D room mesh is typically non-watertight and suffers from severe geometry distortion due to accumulation error brought by the monocular depth estimation method.
%
Moreover, RoomDreamer~\cite{song2023roomdreamer} will jointly refine the geometry and texture of an existing indoor mesh via pretrained text-to-image diffusion models, but still cannot cope with the unobserved regions.
Parallel to scene generation, a line of research works~\cite{yang2023dreamspace,chen2023scenetex,hwang2023text2scene,zhang2023scenewiz3d}, including ours, start paying attention to 3D scene texturing, \ie, generating high-quality textures for given 3D scene-level meshes.
Despite their compelling results, DreamSpace~\cite{yang2023dreamspace} and Text2Scene~\cite{hwang2023text2scene} have to rely on an initial room texture for the succeeding stylization,
while concurrent works SceneTex~\cite{chen2023scenetex} and SceneWiz3D~\cite{zhang2023scenewiz3d} cannot support fine-grained texture controls due to their adopted optimization-based framework.
% Concurrent to this work, DreamSpace~\cite{yang2023dreamspace} and SceneTex~\cite{chen2023scenetex} also propose to synthesize coherent textures for a given scene-level mesh.
% However, Dreamspace requires a mesh scanned from real world as well as an initial room texture and SceneTex cannot support fine-grained controls due to its optimization-based framework.
Unlike any of the above, our model RoomTex targets at texturing indoor scenes that consist of untextured 3D object meshes and simultaneously enables fine-grained controls over the scene.

% !TEX root = ../main.tex

\section{Method}
\label{sec:method}

\begin{figure*}[t!]
	\centering
	\includegraphics[width=\linewidth]{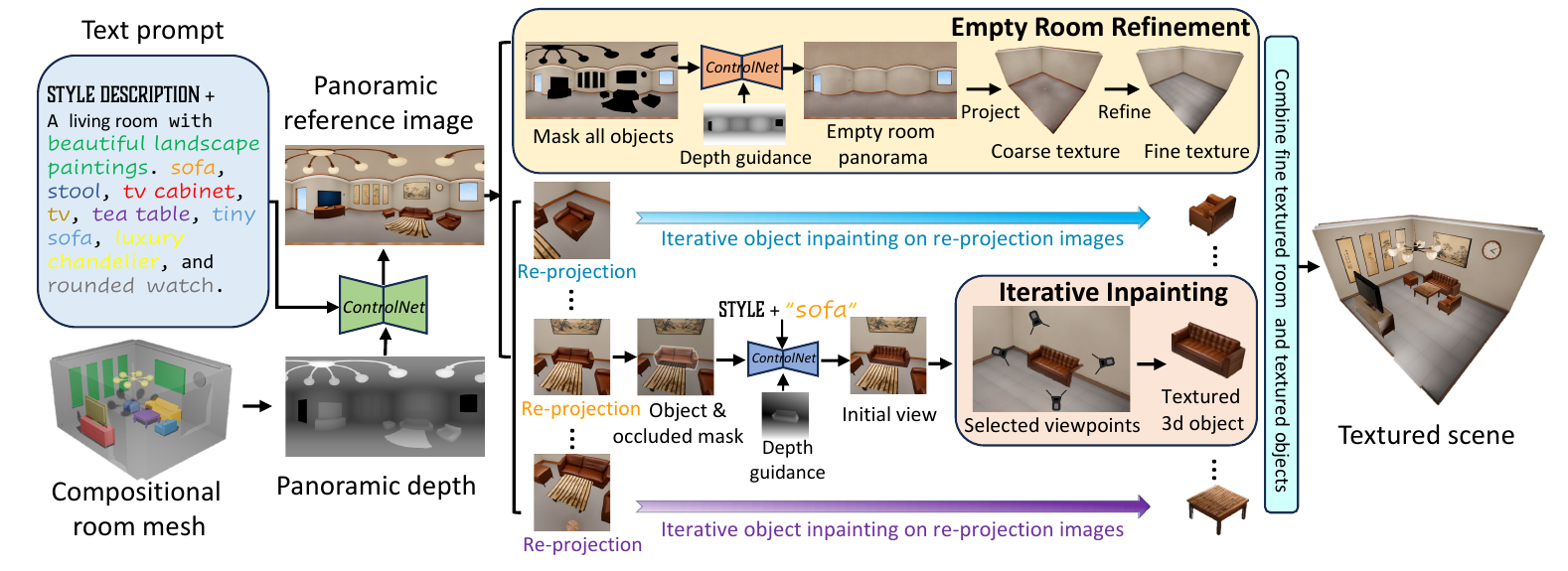}
	\caption{\small \textbf{Framework of RoomTex}. We first generate a panoramic reference image of the indoor scene based on a depth map rendered from a compositional untextured room mesh. Based on the panorama, we will refine and paint every object for a textured 3D object. By integrating objects and the empty room, we can finally get a completely textured 3D indoor scene.}
    \label{fig:method}
    % \vspace{-5pt}
\end{figure*}

In this section, we present our coarse-to-fine generation framework, RoomTex, for synthesizing high-fidelity and style-consistent texture for a compositional room.
We utilize off-the-shelf 3D shape generative models along with a given room layout to assemble the room mesh.
In the coarse stage,
the 3D room mesh is unwrapped to a panorama depth map, based on which we generate a panoramic image of the room as a coarse reference (Sec.~\ref{sec:3.1}).
Then in the fine stage, the empty room will be further refined in perspective views (Sec.~\ref{sec:3.2}).
Afterward,
we employ an iterative inpainting pipeline to refine and paint every independent 3D object in the room (Sec.~\ref{sec:3.3}).
To better align the generated texture and the guidance depth map, we introduce an edge-detection module to identify and then remove the misalignment areas between them (Sec.~\ref{sec:3.4}). Finally, our framework also supports interactive fine-grained texture control (Sec.~\ref{sec:3.5}).
An overview of our framework is illustrated in Fig.~\ref{fig:method}.

%-------------------------------------------------------------------------
\subsection{Panoramic Image Generation}
\label{sec:3.1}

% The initial step of RoomTex is to obtain the room geometry according to a pre-defined room layout and text prompts $\mT$.
% Given text prompts,
% we leverage pretrained 3D shape generative models such as Shap-E~\cite{jun2023shap} to synthesize a collection of object meshes.
% Based on the room layout,
% an empty room mesh can be procedurally generated, which will combine with object meshes and be assembled into a complete 3D room mesh accordingly.
Given the room layout, it is relatively straightforward to assemble an untextured mesh of the room as input by leveraging off-the-shelf 3D object generative models.
Subsequently, a virtual depth camera is put at the center of the room, leading to a panoramic depth map $\mD_p$ of the room via the equirectangular projection.
Under the depth guidance $\mD_p$ and text prompts $\mT$, RoomTex utilizes powerful ControlNet~\cite{zhang2023adding} $\cF_i(\cdot)$ to synthesize a panoramic image $\mI_p$ as a coarse reference to maintain the style consistency of the generated texture:
\begin{equation}\label{eq1}
\mI_p = \cF_i(\mD_p, \mT).
\end{equation}

\subsection{Empty Room Refinement}
\label{sec:3.2}

To enable a more flexible editing to the generated scene like moving furniture items in the scene, we further generate a complete texture of mere interior surfaces using a depth-aware inpainting model so that the missing areas occluded by objects in the initial panorama $\mI_p$ will be filled.
We first remove all the object meshes inside and obtain the panoramic depth of an empty room $\mD_r$.
All the occluded areas are denoted with a binary mask $\mM_r$, and the inpainting can be represented with:
\begin{equation}\label{eq2}
\mI_{r} = \cF_\text{inp}(\mI_p, \mM_r, \mD_r, \mT)
\end{equation}
where $\cF_\text{inp}$ is the depth-aware inpainting model where ControlNet~\cite{zhang2023adding} is used, and the occluded areas in the input pananora $\mI_p$ are assigned zero values during the inpainting.
It is noteworthy that a more complete empty room texture is shown to be beneficial for the subsequent object generation process.

Moreover, to cope with the distortion brought by the panoramic image $\mI_p$, we carefully choose an overhead view targeting the floor and an upward view targeting the ceiling to refine these two important areas.
The panoramic image $\mI_p$ will be re-projected to these two perspective views $\vv_\text{floor}$ and $\vv_\text{ceiling}$ for the corresponding images $\mI_\text{floor}$ and $\mI_\text{ceiling}$ using
\begin{align}\label{eq3}
% \mI_\text{floor} &= \cP(\cT_{\text{pano} \to \text{world}} \circ \mI_p, \vv_\text{floor}), \\
\mI_\cdot &= \cP(\cT_{\text{pano} \to \text{world}} \circ \mI_p, \vv_\cdot),
\end{align}
where $\cT_{\text{pano} \to \text{world}}$ is a transformation function that projects every single pixel in the panoramic image to the spherical coordinates and then projects to the world coordinates with the help of a panoramic depth map,
$\cP(\cdot, \vv)$ is a projection function that projects the point cloud in the world coordinates to a specific view $\vv$.
Regarding $\mI_\text{floor}$ and $\mI_\text{ceiling}$ as the initialization,
we employ the aforementioned inpainting Eq.~\ref{eq2} to refine the floor and ceiling images under the guidance of the corresponding mask and the depth map.
% The same method will be applied to get $\mI_\text{over}$. Once these two images $\mI_\text{up}$ and $\mI_\text{over}$ are acquired, our approach proposes to mask out the ceiling and floor areas and subsequently repaint them:
% \begin{equation}\label{eq4}
% \hat{\mI_\text{up}} = \cF_\text{inp}(\mT, \mD_\text{up}, \mM_\text{up}, \mI_\text{up}, \mM_O)
% \end{equation}
% where $\mD_\text{up}$ is the depth map of the upward view, $\mM_\text{up}$ is the mask of the ceiling area and $\mM_O$ means that the content in the masked area of the initial denoising image is set to be same as the original image $\mI_\text{up}$. The same process will be applied to get a rectified overhead image $\hat{\mI_\text{over}}$ targeting the floor. In the end of this stage, the masked area in $\hat{\mI_\text{over}}$ and $\hat{\mI_\text{up}}$ will be re-projected back to the panoramic image $\mI_r$ with mere rectified room interior surfaces.

%-------------------------------------------------------------------------
\subsection{Iterative Object Texturing}
\label{sec:3.3}
Notably, generating a panoramic image is not essentially equal to generating a complete room texture supporting free novel view rendering inside mainly due to the lack of information in the occluded area.
Therefore, it is necessary to apply an inpainting process to fill in the `other' side of every single object, \ie, the missing areas in the panorama.
In this stage, the global panorama $\mI_p$ and the empty room panorama $\mI_r$ are further used as references for the texturing of each object in the scene.
Our method RoomTex aims to generate texture for a compositional 3D scene and thus conduct the inpainting for each independent object separately.

For a specific 3D object to be painted,
we will select an initial perspective view $\vv_0$ targeting it and re-project $\mI_p$ and $\mI_r$ to obtain foreground and background images $\mI^\text{fg}_\text{obj}$ and $\mI^\text{bg}_\text{obj}$ of this object via Eq.~\ref{eq3}.
In particular, the perspective image resolution is set as a constant, and the focal length of the initial view $\vv_0$ is related to the size of the specific object and the relative distance between the camera and the object.
With perspective images $\mI_\text{bg}$ and $\mI_\text{fg}$, we further integrate them into one image $\hat{\mI_\text{obj}}$ using the object mask $\mM_\text{obj}$ as follows:
\begin{equation}\label{eq4}
\hat{\mI_\text{obj}} = \mI_\text{fg} \odot \mM_\text{obj} + \mI_\text{bg} \odot (1 - \mM_\text{obj}).
\end{equation}
Afterward, this fused image $\hat{\mI_\text{obj}}$ is refined to a new image $\mI_\text{obj}$ with less distortion and higher resolution via Eq.~\ref{eq2}.

\begin{figure*}[t!]
	\centering
	\includegraphics[width=0.97\linewidth]{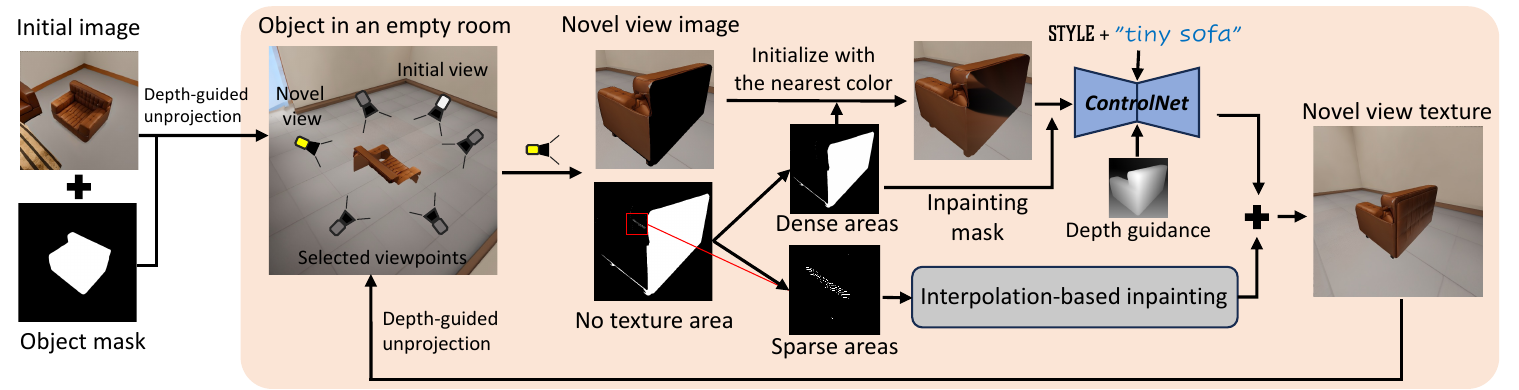}
	\caption{\small \textbf{Iterative inpainting.} We leverage the object depth to unproject only object areas of the initial image to the world coordinates. Then, we choose a group of suitable views and iteratively warp the 3D object to these views, under which the untextured area will be filled with diffusion-based inpainting (dense areas) and interpolation-based inpainting (sparse areas).
 }
    \label{fig:iterative}
\end{figure*}

After obtaining the refined initial view $\mI_\text{obj}$ of the object,
we leverage iterative inpainting to completely texture this object as illustrated in Fig.~\ref{fig:iterative}.
To this end, we extra select a series of camera viewpoints $\{ \vv_i \}, i=1,2,...,N$,
which will cover every aspect of the object as comprehensively as possible.
The first eight views place the camera on a sphere centered around the object looking at the center of the object.
To be specific, the radius of the sphere is set to be slightly larger than the object and the polar angle is set to $\pi/4$ and $3\pi/4$.
Moreover,
some additional views will be picked if this 3D object is out of the view range of eight selected camera poses.
Once the group of views $\left\{\vv_i\right\}$ is acquired,
we can apply an iterative warping and inpainting process to texture this 3D object.
First of all, the initial image is unprojected to partial point cloud $\mP_\text{obj}$ in the world coordinate under the depth guidance:
\begin{equation}\label{eq7}
\mP_\text{obj} =  \cP^{-1} (\mI_\text{obj} \odot \mM_\text{obj}, \vv_0),
\end{equation}
where $\cP^{-1} (\cdot, \vv)$ aims to unproject pixels in the perspective view $\vv$ back to the world coordinate.

For a novel view $\vv_i$,
the partial point cloud $\mP_\text{obj}$ will be warped to a novel view image $\hat{\mI_\text{obj, i}}$ to be inpainted via Eq.~\ref{eq4},
and the corresponding inpainting mask $\mM_\text{obj, i}$ can be also obtained simultaneously.
During the inpainting,
we will initialize the missing areas with the nearest neighbor color to guide the inpainting model for a style-consistent texture,
and then inpaint the image $\hat{\mI_\text{obj, i}}$ to $\mI_\text{obj, i}$ via aforementioned Eq.~\ref{eq2}.
However,
this inpainted image $\mI_\text{obj, i}$ inevitably contains numerous sparse and small holes due to the sparsity of partial point cloud $\mP_\text{obj}$, which cannot be filled well with diffusion-based inpainting models like ControlNet~\cite{zhang2023adding}.
Therefore,
we extra leverage an interpolation-based inpainting method $\cF_\text{intp}$ to fill these sparse areas indicated by a binary mask $\mM^s_\text{obj, i}$.
Combining with the room background,
the final image under this novel view $\mI^\text{final}_\text{obj, i}$ can be represented as follows:
\begin{equation} \label{eq-combine}
\begin{split}
    \mI^\text{final}_\text{obj, i} &= \hat{\mI_\text{obj, i}} \odot (1 - \mM_\text{obj, i}) +  \cF_\text{intp}(\hat{\mI_\text{obj, i}}) \odot \mM^s_\text{obj, i} \\
    &\quad + \mI_\text{obj, i} \odot (\mM_\text{obj, i} - \mM^s_\text{obj, i}).
\end{split}
\end{equation}
Afterward,
the final image in this novel view will be unprojected to the world coordinate under the depth guidance and then merge with the previous partial point cloud $\mP_\text{obj}$ following:
\begin{equation}\label{eq-pc}
\mP_\text{obj} :=  \mP_\text{obj} \cup (\cP^{-1} (\mI^\text{final}_\text{obj, i} \odot \mM_\text{obj, i}, \vv_i)),
\end{equation}
where the updated point cloud will engage in the next iteration of object texturing, and $\cup$ denotes the set union operation. 
The iterative inpainting will be conducted following our selected camera poses $\{ \vv_i \}, i=1,2,...,N$,
until the 3D object is completely textured.

\begin{figure}[t!]
	\centering
	\begin{minipage}{0.48\linewidth}
		\centering
		\includegraphics[width=\linewidth]{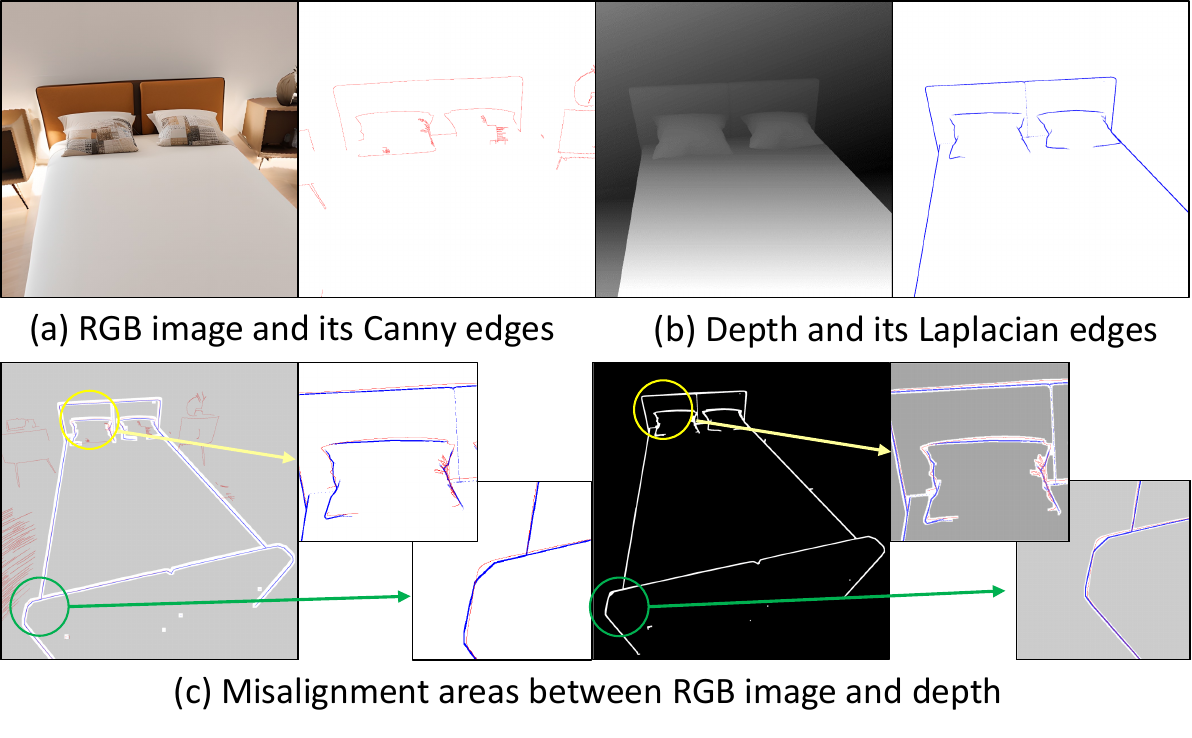}
            \vspace{-18pt}
		\caption{\small \textbf{Misalignment removal.} We first get the Canny edges of RGB images and Laplacian edges of depth maps as shown in (a) and (b). (c) shows the misalignment areas between texture and depth, which will be removed during the unprojection.}
            \label{fig:misalignment}
	\end{minipage}
	\hfill
	\begin{minipage}{0.48\linewidth}
		\centering
            \vspace{-3pt}
		\includegraphics[width=\linewidth]{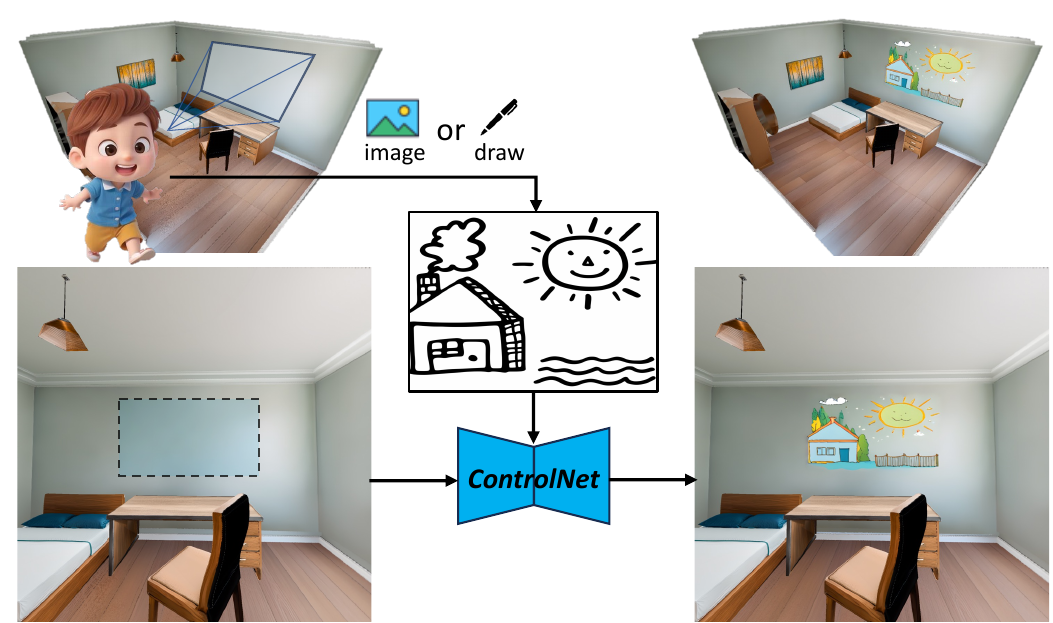}
		\caption{
			\small{\textbf{Fine-grained texture control.} Users could interactively point out the specific area they would like to edit together with a sketch illustrating how they would like to edit this area. We take the sketch as an additional input to ControlNet to achieve fine-grained control.}
		}
		\label{fig:control}
	\end{minipage}
	\vspace{0pt}
\end{figure}

% \begin{figure}[t!]
% 	\centering
% 	\includegraphics[width=0.98\linewidth]{figs/misalignment.pdf}
% 	\caption{\small \textbf{Misalignment removal.} We first get the Canny edges of RGB images and Laplacian edges of depth maps as shown in (a) and (b). (c) shows the misalignment areas between texture and depth, which will be removed during the unprojection.}
%     \label{fig:misalignment}
%     \vspace{-5pt}
% \end{figure}

%-------------------------------------------------------------------------
\subsection{Misalignment Removal}
\label{sec:3.4}
As illustrated in Fig.~\ref{fig:misalignment}, the generated texture cannot perfectly align with the depth map, especially in the areas around the depth edge.
To mitigate this problem, a carefully designed edge detection module is proposed to identify the misalignment areas.
We denote a perspective image from diffusion models as $\mI$, its Canny edges as $\cE_{C}(\mI)$, depth map as $\mD$, and its Laplacian edges as $\cE_{L}(\mD)$.
Then, our method will leverage traditional erosion and dilatation operations to determine the misalignment areas.
Specifically,
we filter out all the irrelevant areas by dilating the Laplacian edges and only keep the overlapping part:
\begin{equation}\label{eq-dilate}
\hat{\cE_{C}(\mI)} = \cE_{C}(\mI) \cap \text{Dilate}(\cE_{L}(\mD)).
\end{equation}
Afterward, the mask of the misaligned area can be obtained via erosion and dilatation operations:
\begin{equation}\label{eq-final}
\mM_\text{mis} = \text{Erode} \big(\text{Dilate} (\hat{\cE_{C}(\mI)} \cup \cE_{L} (\mD)) \big).
\end{equation}
Misalignment areas will not be taken into consideration during the unprojection.

\subsection{Fine-grained Texture Control}
\label{sec:3.5}
In practice,
users may not be completely satisfied with the generated textures and would like to rectify specific areas, or even provide more detailed controls to ensure the generated textures meet their expectations as illustrated in Fig.~\ref{fig:control}.
We denote the image from the specific viewpoint $\vv$ that the user wishes to edit as $\mI$, the area they want to interact as $\mM$, the corresponding depth map as $\mD$, and a sketch illustrating how they would like to edit this area as $\mS$. Then, we will repaint the masked-out area according to the additional sketch conditions:
\begin{equation}\label{eq-control}
\mI^{\prime} = \cF_s(\mI, \mM, \mD, \mS, \mT)
\end{equation}
Afterward, the newly painted area will be projected back to the world coordinates, updating the point cloud of the scene $\mP_\text{scene}$:
\begin{equation}\label{eq-pc-fine}
\mP_\text{scene} :=  \mP_\text{scene} \setminus \mP_\text{orig} \cup (\cP^{-1} (\mI^{\prime} \odot \mM, \vv)),
\end{equation}
where $\mP_\text{orig}$ denotes the original point cloud of the masked area and $\setminus$ stands for set subtraction operation.
It is worth noting that users may generate their desirable sketches using text-to-image models. Moreover, if users just want to repaint several unsatisfactory objects, our methods also support object-level texture control like changing the color with mere text guidance as shown in Fig.\ref{fig:teaser-control}.
% !TEX root = ../main.tex

\section{Experiment}
\label{sec:exp}
In this section, we will present qualitative and quantitative results as well as our ablation study.
For implementation details, please refer to the supplemental material, and we will first give a brief introduction to the baseline methods.

% \noindent\textbf{Implementation Details.} 
% The furniture models placed in our scene are generated by ourselves using simple text prompts and Shap-E\cite{jun2023shap}. All the room sketch layouts for different kinds of rooms in the experiment is designed by ourselves, which isn't a hard process and users can design the layout of their rooms as well. We utilize a Stable Diffusion ~\cite{rombach2022stablediffusion} and a depth-guided ControlNet\cite{zhang2023adding} model for $\cF_i$, and the model is further finetuned on the image inpainting task for $\cF_\text{inp}$. Normally, the default value of the focal length of our perspective camera is $f=500$, which may be adjusted if the object is too far away from the camera. The threshold value for Canny Edge Detector is set to 200 and 800 in all our experiments. Furthermore, we leverage a Fluid Dynamics Method in OpenCV as our interpolation-based inpainting method. All of our experiments are conducted on NVIDIA A100(80G) GPUs, and generating a scene takes approximately 90 minutes on multiple GPUs depending on the number of furniture items in the scene.

\begin{figure*}[t!]
    \centering
    \includegraphics[width=0.95\linewidth]{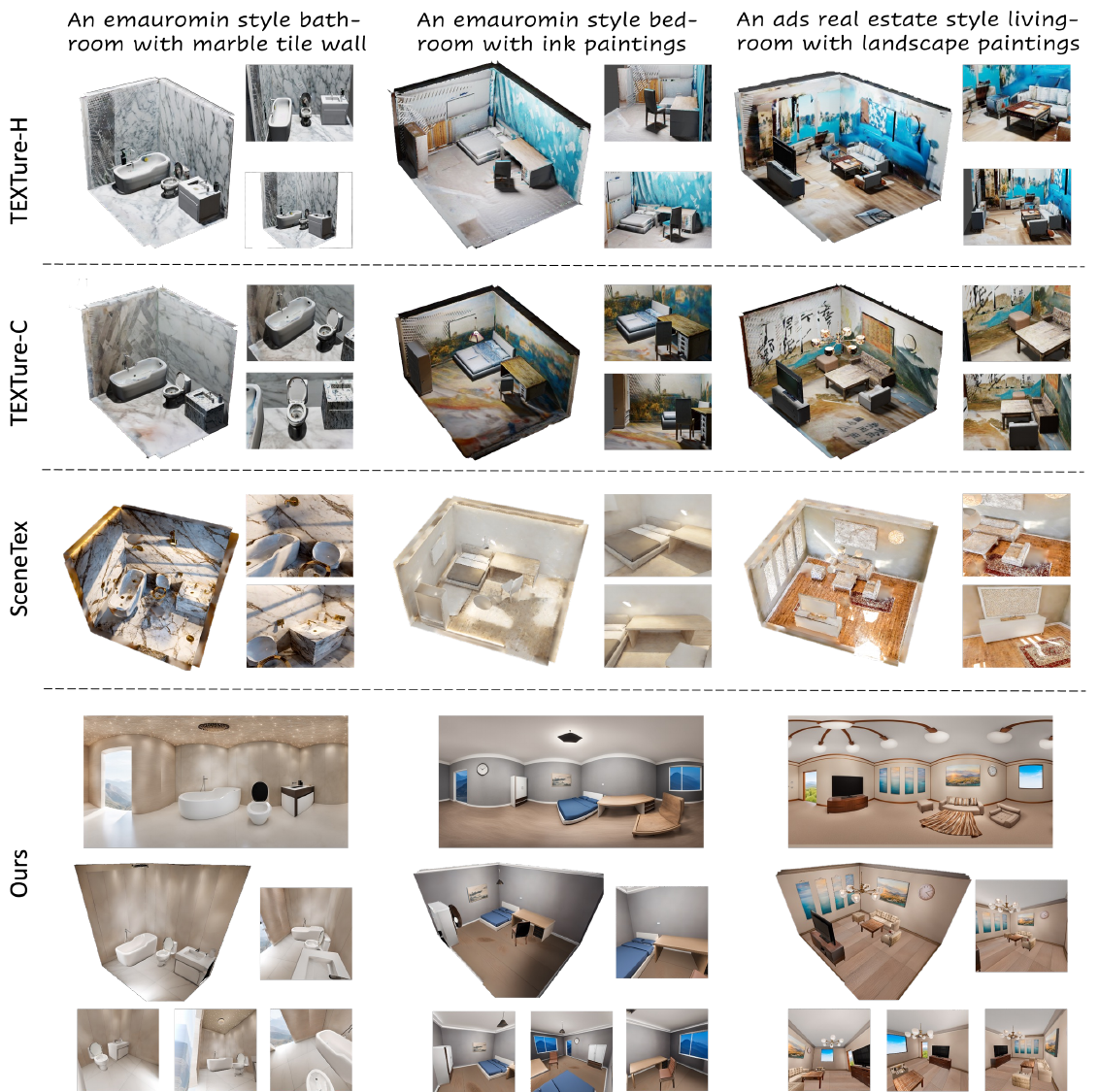}
    % \vspace{-7pt}
    \caption{\small We compare our generated textured scene with TEXTure\cite{richardson2023texture} and SceneTex\cite{chen2023scenetex}, where the figures include an overhead view and several views rendered from inside of the scene. Our reference panorama is also shown. (\textbf{Zoom in for best view})}
    \label{fig:comparison}
    \vspace{-7pt}
\end{figure*}

\begin{figure*}[t!]
    \centering
    \includegraphics[width=0.95\linewidth]{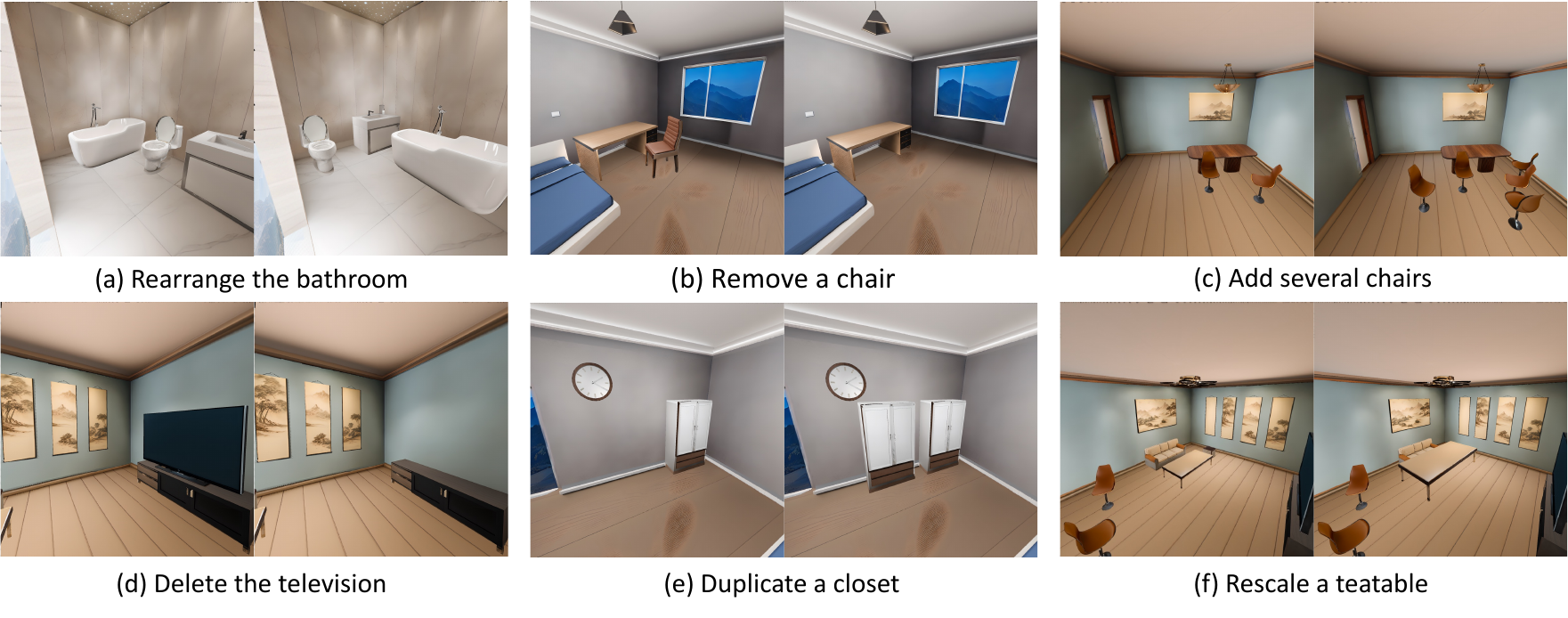}
    \vspace{-7pt}
    \caption{\small \textbf{Scene editing.} Here we show how the compositional design empowers flexible editing. Rearranging, removing, adding, deleting, duplicating, and rescaling objects in different scenes are naturally supported.}
    \label{fig:scene-editing}
    \vspace{-10pt}
\end{figure*}

\begin{figure}
    \includegraphics[width=\linewidth]{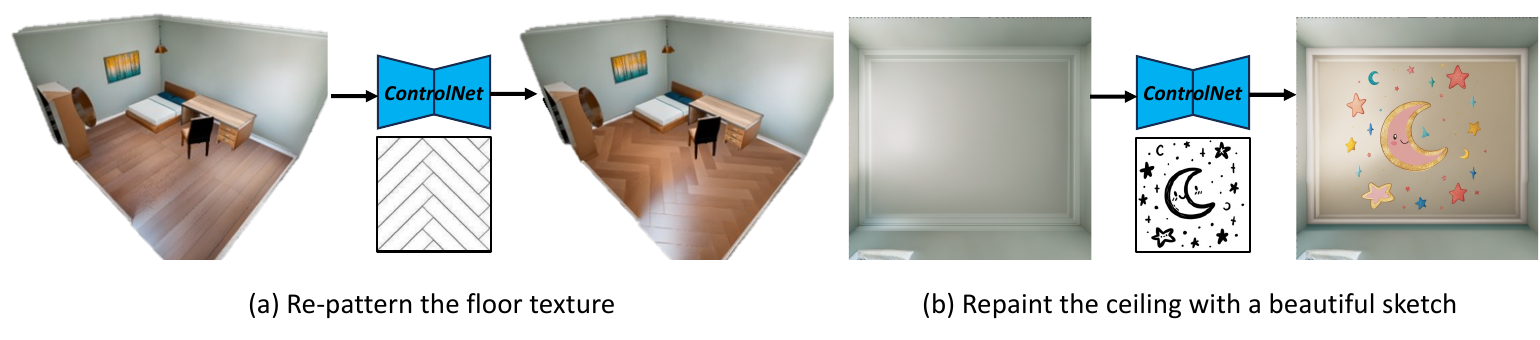}
    \vspace{-16pt}
    \caption{\small \textbf{Fine-grained texture control results.} We show two examples of interactive fine-grained texture control. On the left, we show a top view comparison of the floor before and after re-patterning the floor texture. On the right, we show an upward view comparison of the ceiling before and after repainting the ceiling texture.}
    \vspace{-8pt}
    \label{fig:fine-grain-results}
\end{figure}

\noindent\textbf{Baselines.}
We compare our method against two recent texture synthesis methods. As for MVDiffusion~\cite{tang2023mvdiffusion}, we found the pre-trained depth-guided holistic generation model only works well for ScanNet~\cite{dai2017scannet} and can only generate blur images that do not align well with the depth guidance in cases for comparisons.
\begin{itemize}
% \item[$\bullet$] Text2Room~\cite{hollein2023text2room}: We compare with Text2Room whose input is mere text descriptions. We utilize the same text description as ours to guide the generation process.

\item[$\bullet$]TEXTure-H~\cite{richardson2023texture}: We compare with TEXTure whose input is an untextured mesh and text prompts. We holistically implement this baseline by removing several walls and the ceiling in our scenes so that the texture of interior surfaces can be generated.
\item[$\bullet$]TEXTure-C\cite{richardson2023texture}: We implement this baseline by compositionally generating the texture of every object mesh and the room interior surfaces, whose several walls and ceilings will also be removed. Afterward, we would integrate all components into a holistic mesh. 
\item[$\bullet$]SceneTex~\cite{chen2023scenetex}: We compare with SceneTex whose input is also an untextured mesh and text prompts. We utilize the exact same text description as ours to guide the generation process.
\end{itemize}

\noindent\textbf{Evaluation Metrics.} The generated 3D textured room is evaluated both quantitatively and qualitatively. We leverage the Aesthetic Score(AS) introduced by LAION~\cite{schuhmann2021laion}, CLIP Score(CS)~\cite{radford2021learning} and BRISQUE(BQ)~\cite{mittal2012no} to reflect the image quality of the generated scenes. 
% Additionally, we conduct a user study and ask n=xx users to score 3D consistency(3DC), Texture Quality(TQ), and Perceptual Quality(PQ) on the whole scene on a scale of 1-5. 

\subsection{Qualitative Results}

\noindent\textbf{Comparison to baselines.}
For comparison with baseline methods, we generate three scenes including a bathroom, a bedroom, and a living room. We show top-down views into the scene and several perspective images for our method and baselines in Fig.~\ref{fig:comparison}.
% TEXTure-W can texture a room mesh, however, the texture of the interior surfaces is far from satisfactory and there may exist unnatural texture in specific areas of an object because it's hard to describe every aspect of the scene with the same text. The textured scene generated by TEXTure-C can also support flexible editing, however, the texture of interior surfaces is still imperfect and the texture between different objects aren't harmonious for lack of a holistic reference. Both TEXTure-C and TEXTure-W will generate some unnatural spots and stripes. 
Neither TEXTure-H nor TEXTure-C can generate satisfactory interior surfaces, and unnatural spots and stripes may appear on the individual objects.
Furthermore, the texture generated by TEXTure-C is inconsistent for lack of a holistic constraint and the texture generated by TEXTure-H is also of low quality due to the inaccurate description.
% SceneTex can generate a relatively consistent texture for the whole scene including interior surfaces. However, the generated texture does not perfectly align with the given geometry. For example, strange circle-like objects appear on the walls of the bathroom and a strange box-like object appears in front of the bed in the bedroom. Moreover, the generated texture contains lots of strange noise and unreasonable lighting, and the texture occluded due to viewpoint choice seems blurry.
SceneTex~\cite{chen2023scenetex} can generate a relatively consistent texture for the whole scene including interior surfaces.
However, the generated texture contains lots of unsatisfactory noise, unreasonable lighting, and severe misalignment between texture and geometry.
For example, strange circle-like objects appear on the walls of the bathroom and a strange box-like object appears in front of the bed in the bedroom.
More crucially, SceneTex cannot thoroughly settle the occlusion problem in the scene, resulting in blurry texture in the occluded areas.
In contrast, our approach creates a highly detailed and compelling texture for all the 3D objects in the scene.
Thanks to the panorama reference, the texture style between different objects and the context (mostly walls and the floor) is coherent.
Moreover, the style of the room can be easily modified as shown in Fig.~\ref{fig:teaser}.

\noindent\textbf{Scene editing and Fine-grained texture control.}
Our texture is generated for a compositional scene including objects and surroundings, and the representation of every object is essentially a colored point cloud. Therefore, editing like adding, duplicating, removing, rotating, moving, and rescaling objects can be simply supported.
In Fig.~\ref{fig:scene-editing}, we include several editing results of rendered images under various views in different scenes.
Moreover, as illustrated in Sec.~\ref{sec:3.5}, interactive fine-grained control over texture can be easily achieved and two additional examples are provided in Fig.~\ref{fig:fine-grain-results}.

% \noindent\textbf{Fine-grained control.}
% Since our entire texturing framework is explicit and some fine-grained controls can be achieved through sketches together with ControlNet~\cite{zhang2023adding}, we can go beyond just using depth guidance. As shown in Fig.~\ref{fig:control}, users may want to add specific patterns to a certain area. They may either acquire the sketch with the help of text-to-image models or draw the sketch themselves. Afterward, the certain area will be repainted under the guidance of the sketch to generate more controllable texture. More examples are shown in Fig.~\ref{fig:fine-grain}.

\subsection{Quantitative Results}
We show quantitative results averaged over multiple scenes including a bathroom, a bedroom, and a living room in Tab.\ref{tab:userstudy}.
We render about 100 images from novel views for each scene to calculate these three metrics. 
Blurry and messy texture lead to lower scores for the baselines in image-based scores.
As for the computing of CS, it is hard to offer an accurate description for all rendered images, so the general text input is used instead.
TEXTure-C leverages the text prompt for every object inside, thus leading to a higher CS score than ours.
SceneTex is prone to generating evident lighting, leading to a higher BQ value under specific viewpoints since BQ metric favors images with evident lighting.
% which explains why the score of TEXTure-C is a little higher than ours. A user study is conducted as well. Please refer to the supplemental material for details.
% In user study, we present users with a room tour video for each scene and let them rate the generation quality of each method. 
% Moreover, users prefer our method, which highlights the high quality of our generated texture and the harmonious texture style of the holistic scene.

\begin{table}[!t]
\caption{
\textbf{Quantitative comparison.} We report image quality metrics including Aesthetic Score(AS), Clip Score(CS), BRISQUE(BQ). Our method outperforms baselines on AS, slightly worse than TEXTure-C in terms of CS and slightly worse than SceneTex in terms of BQ.
}
\vspace{-17pt}
\begin{center}
\footnotesize
\begin{tabular}{lccccc}
\hline
Method &AS($\uparrow$) &CS($\uparrow$) &BQ($\downarrow$) \\
\hline
TEXTure-C~\cite{richardson2023texture} & 5.11 & \textbf{30.17} & 39.78\\ 
TEXTure-H~\cite{richardson2023texture} & 4.33 & 27.86 & 47.54\\
SceneTex~\cite{chen2023scenetex}  & 4.77 & 26.43 & \textbf{26.91}\\ 
Ours & \textbf{5.20} & 29.16 & 30.91 \\ 
\hline
\end{tabular}

\end{center}
% \caption{
% \textbf{Quantitative comparison.} We report image quality metrics including Aesthetic Score(AS), Clip Score(CS), BRISQUE(BQ). Our method outperforms baselines on AS, slightly worse than TEXTure-C in terms of CS and slightly worse than SceneTex in terms of BQ.
% }
\vspace{-17pt}
% \vspace{-10pt}

\label{tab:userstudy}
\end{table}

\subsection{Ablation Study}
% The key ingredients of our method are depth-guided projection mask(Sec.\ref{sec:3.3}), equirectangular projection distortion elimination(Sec.\ref{sec:3.1}), and initialization of denoising image(Sec.\ref{sec:3.2}).

\begin{figure}[t!]
    \centering
    \includegraphics[width=0.95\linewidth]{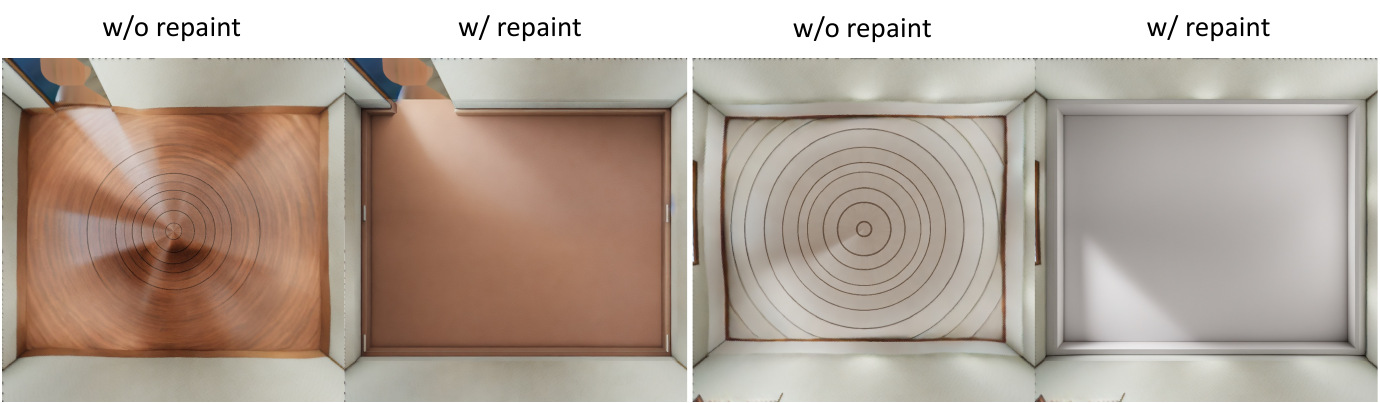}
    % \vspace{-12pt}
    \caption{\small \textbf{Ablation study on distortion elimination.} Repainting results of floors and ceilings are significantly better than those without repaint.}
    \vspace{-4pt}
    \label{fig:ablation-repaint}
\end{figure}

\noindent\textbf{Panorama distortion elimination.} It is noteworthy that except for the object distortion brought by the equirectangular projection, the texture of the ceilings, floor, and baseboards may also suffer from distortion as shown in Fig.~\ref{fig:ablation-repaint}.
This kind of texture is unacceptable and may further influence the iterative object inpainting process since the room context (background texture) is unrealistic.
After choosing an overhead and an upward view to repaint floors and ceilings, we can get a textured empty room with less distortion.

\begin{figure}[t!]
	\centering
	\begin{minipage}{0.47\linewidth}
		\centering
            % \vspace{-3pt}
            \includegraphics[width=\linewidth]{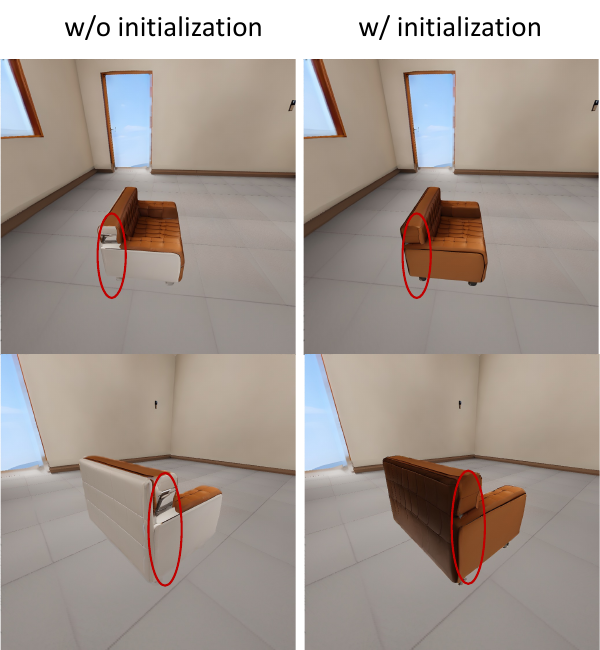}
	    \caption{\small \textbf{Ablation study on initialization of untextured areas.} We show generated texture without initializing the untextured areas, where messy texture and a large inharmonious white area appear. In contrast, the initialization can mitigate this problem.}
            \label{fig:ablation-initialization}
	\end{minipage}
	\hfill
	\begin{minipage}{0.5\linewidth}
		\centering
            %\vspace{-3pt}
		\includegraphics[width=\linewidth]{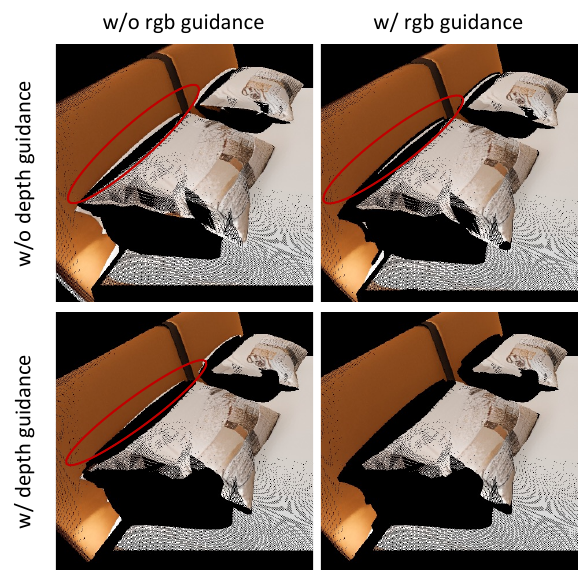}
            %\vspace{-18pt}
		\caption{\small \textbf{Ablation study on misalignment detection.} We show results of using different misalignment detection techniques, including using no guidance, mere RGB guidance, mere depth guidance, and both. Only using both introduced in our method can avoid the gray texture on the pillow dilating to the headboard.}
            \label{fig:ablation-misalignment}
	\end{minipage}
	% \vspace{-10pt}
\end{figure}

\noindent\textbf{Initialization of untextured areas.} In Sec.~\ref{sec:3.3}, initialization of the untextured areas may contribute to a higher quality of inpainting if the missing texture is of a similar color to its nearby area.
As shown in Fig.~\ref{fig:ablation-initialization}, whether or not we fill the untextured areas with the nearest color will make a huge difference in the generation quality.

% \begin{figure}
% 	\includegraphics[width=0.9\linewidth]{figs/scene-editing.pdf}
% 	\caption{\small Scene editing.}
%     \label{fig:scene-editing}
% \end{figure}
% It's noteworthy that except for the object distortion brought by the equirectangular camera, the texture of the ceilings, floor and baseboards also suffer from distortion as shown in Fig.~\ref{fig:ablation-repaint}. This kind of texture is far from satisfactory and may further influence the object inpainting process since the context (background texture) is a little messy. After carefully choosing an overhead view and an upward view to repaint these areas, we can finally get an interior-surfaces panorama with less distortion.

\noindent\textbf{Misalignment detection.}
As mentioned in Sec.~\ref{sec:3.4}, misalignment between the generated texture and the guiding depth map often occurs, which will dilate the foreground texture to the background as shown in the top-left image in Fig.~\ref{fig:ablation-misalignment}.
For example, the texture of the pillows may dilate into the area of the headboard, which will interfere with the inpainting model in the later iteration since the texture warped from previous frames is not convincing.
These areas will be detected and then discarded during the unprojection.
Visualized reprojection results of using different kinds of projection masks are shown from a new viewpoint.

% !TEX root = ../main.tex

\section{Conclusion and Limitation}
We have proposed a novel text-driven indoor scene texture generation framework, which is capable of generating high-fidelity and coherent texture that aligns well with geometry.
The crux of our approach is to first synthesize a panoramic image as a holistic reference for style consistency and then inpaint every object iteratively to support a compositional 3D scene with complete and harmonious textures.
Experimental results demonstrate the superiority of our approach concerning generation quality and editing flexibility.
\\
\textbf{Limitations and future work.}
% Our framework still relies on other pretrained 3D shape generative models.
Our iterative inpainting strategy is incapable of capturing all views of a 3D object in one run, potentially leading to inconsistent texture despite all the refinement we have applied.
We believe multi-view diffusion models~\cite{shi2023mvdream, tang2023mvdiffusion} may mitigate this problem.

% ---- Bibliography ----
%
% BibTeX users should specify bibliography style 'splncs04'.
% References will then be sorted and formatted in the correct style.
%
\bibliographystyle{splncs04}
\bibliography{main}

\begin{thebibliography}{10}
\providecommand{\url}[1]{\texttt{#1}}
\providecommand{\urlprefix}{URL }
\providecommand{\doi}[1]{https://doi.org/#1}

\bibitem{balaji2022ediffi}
Balaji, Y., Nah, S., Huang, X., Vahdat, A., Song, J., Kreis, K., Aittala, M., Aila, T., Laine, S., Catanzaro, B., et~al.: ediffi: Text-to-image diffusion models with an ensemble of expert denoisers. arXiv preprint arXiv:2211.01324  (2022)

\bibitem{bokhovkin2023mesh2tex}
Bokhovkin, A., Tulsiani, S., Dai, A.: Mesh2tex: Generating mesh textures from image queries. arXiv preprint arXiv:2304.05868  (2023)

\bibitem{canny1986computational}
Canny, J.: A computational approach to edge detection. IEEE Transactions on pattern analysis and machine intelligence (6),  679--698 (1986)

\bibitem{cao2023texfusion}
Cao, T., Kreis, K., Fidler, S., Sharp, N., Yin, K.: Texfusion: Synthesizing 3d textures with text-guided image diffusion models. In: Proceedings of the IEEE/CVF International Conference on Computer Vision. pp. 4169--4181 (2023)

\bibitem{shapenet2015}
Chang, A.X., Funkhouser, T., Guibas, L., Hanrahan, P., Huang, Q., Li, Z., Savarese, S., Savva, M., Song, S., Su, H., Xiao, J., Yi, L., Yu, F.: {ShapeNet: An Information-Rich 3D Model Repository}. Tech. Rep. arXiv:1512.03012 [cs.GR], Stanford University --- Princeton University --- Toyota Technological Institute at Chicago (2015)

\bibitem{chen2023scenetex}
Chen, D.Z., Li, H., Lee, H.Y., Tulyakov, S., Nie{\ss}ner, M.: Scenetex: High-quality texture synthesis for indoor scenes via diffusion priors. arXiv preprint arXiv:2311.17261  (2023)

\bibitem{chen2023text2tex}
Chen, D.Z., Siddiqui, Y., Lee, H.Y., Tulyakov, S., Nie{\ss}ner, M.: Text2tex: Text-driven texture synthesis via diffusion models. arXiv preprint arXiv:2303.11396  (2023)

\bibitem{chen2023fantasia3d}
Chen, R., Chen, Y., Jiao, N., Jia, K.: Fantasia3d: Disentangling geometry and appearance for high-quality text-to-3d content creation. arXiv preprint arXiv:2303.13873  (2023)

\bibitem{chen2022tango}
Chen, Y., Chen, R., Lei, J., Zhang, Y., Jia, K.: Tango: Text-driven photorealistic and robust 3{d} stylization via lighting decomposition. NeurIPS  (2022)

\bibitem{chen2022auv}
Chen, Z., Yin, K., Fidler, S.: Auv-net: Learning aligned uv maps for texture transfer and synthesis. In: Proceedings of the IEEE/CVF Conference on Computer Vision and Pattern Recognition. pp. 1465--1474 (2022)

\bibitem{cohen2023set}
Cohen-Bar, D., Richardson, E., Metzer, G., Giryes, R., Cohen-Or, D.: Set-the-scene: Global-local training for generating controllable nerf scenes. arXiv preprint arXiv:2303.13450  (2023)

\bibitem{dai2017scannet}
Dai, A., Chang, A.X., Savva, M., Halber, M., Funkhouser, T., Nie{\ss}ner, M.: Scannet: Richly-annotated 3d reconstructions of indoor scenes. In: Proceedings of the IEEE conference on computer vision and pattern recognition. pp. 5828--5839 (2017)

\bibitem{deitke2023objaverse}
Deitke, M., Schwenk, D., Salvador, J., Weihs, L., Michel, O., VanderBilt, E., Schmidt, L., Ehsani, K., Kembhavi, A., Farhadi, A.: Objaverse: A universe of annotated 3d objects. In: Proceedings of the IEEE/CVF Conference on Computer Vision and Pattern Recognition. pp. 13142--13153 (2023)

\bibitem{fang2023ctrl}
Fang, C., Hu, X., Luo, K., Tan, P.: Ctrl-room: Controllable text-to-3d room meshes generation with layout constraints. arXiv preprint arXiv:2310.03602  (2023)

\bibitem{fridman2023scenescape}
Fridman, R., Abecasis, A., Kasten, Y., Dekel, T.: Scenescape: Text-driven consistent scene generation. arXiv preprint arXiv:2302.01133  (2023)

\bibitem{fu20213d}
Fu, H., Cai, B., Gao, L., Zhang, L.X., Wang, J., Li, C., Zeng, Q., Sun, C., Jia, R., Zhao, B., et~al.: 3d-front: 3d furnished rooms with layouts and semantics. In: Proceedings of the IEEE/CVF International Conference on Computer Vision. pp. 10933--10942 (2021)

\bibitem{gao2022get3d}
Gao, J., Shen, T., Wang, Z., Chen, W., Yin, K., Li, D., Litany, O., Gojcic, Z., Fidler, S.: Get3d: A generative model of high quality 3d textured shapes learned from images. Advances In Neural Information Processing Systems  \textbf{35},  31841--31854 (2022)

\bibitem{gupta20233dgen}
Gupta, A., Xiong, W., Nie, Y., Jones, I., O{\u{g}}uz, B.: 3dgen: Triplane latent diffusion for textured mesh generation. arXiv preprint arXiv:2303.05371  (2023)

\bibitem{ho2020denoising}
Ho, J., Jain, A., Abbeel, P.: Denoising diffusion probabilistic models. NeurIPS  \textbf{33},  6840--6851 (2020)

\bibitem{hollein2023text2room}
H{\"o}llein, L., Cao, A., Owens, A., Johnson, J., Nie{\ss}ner, M.: Text2room: Extracting textured 3d meshes from 2d text-to-image models. arXiv preprint arXiv:2303.11989  (2023)

\bibitem{hwang2023text2scene}
Hwang, I., Kim, H., Kim, Y.M.: Text2scene: Text-driven indoor scene stylization with part-aware details. In: Proceedings of the IEEE/CVF Conference on Computer Vision and Pattern Recognition. pp. 1890--1899 (2023)

\bibitem{jun2023shap}
Jun, H., Nichol, A.: Shap-e: Generating conditional 3d implicit functions. arXiv preprint arXiv:2305.02463  (2023)

\bibitem{li2023sweetdreamer}
Li, W., Chen, R., Chen, X., Tan, P.: Sweetdreamer: Aligning geometric priors in 2d diffusion for consistent text-to-3d. arXiv preprint arXiv:2310.02596  (2023)

\bibitem{lin2023magic3d}
Lin, C.H., Gao, J., Tang, L., Takikawa, T., Zeng, X., Huang, X., Kreis, K., Fidler, S., Liu, M.Y., Lin, T.Y.: Magic3d: High-resolution text-to-3d content creation. In: Proceedings of the IEEE/CVF Conference on Computer Vision and Pattern Recognition. pp. 300--309 (2023)

\bibitem{liu2023unidream}
Liu, Z., Li, Y., Lin, Y., Yu, X., Peng, S., Cao, Y.P., Qi, X., Huang, X., Liang, D., Ouyang, W.: Unidream: Unifying diffusion priors for relightable text-to-3d generation. arXiv preprint arXiv:2312.08754  (2023)

\bibitem{liu2023meshdiffusion}
Liu, Z., Feng, Y., Black, M.J., Nowrouzezahrai, D., Paull, L., Liu, W.: Meshdiffusion: Score-based generative 3d mesh modeling. arXiv preprint arXiv:2303.08133  (2023)

\bibitem{metzer2023latent}
Metzer, G., Richardson, E., Patashnik, O., Giryes, R., Cohen-Or, D.: Latent-nerf for shape-guided generation of 3d shapes and textures. In: Proceedings of the IEEE/CVF Conference on Computer Vision and Pattern Recognition. pp. 12663--12673 (2023)

\bibitem{text2mesh}
Michel, O., Bar-On, R., Liu, R., Benaim, S., Hanocka, R.: Text2mesh: Text-driven neural stylization for meshes. CVPR  (2022)

\bibitem{michel2022text2mesh}
Michel, O., Bar-On, R., Liu, R., Benaim, S., Hanocka, R.: Text2mesh: Text-driven neural stylization for meshes. In: Proceedings of the IEEE/CVF Conference on Computer Vision and Pattern Recognition. pp. 13492--13502 (2022)

\bibitem{mittal2012no}
Mittal, A., Moorthy, A.K., Bovik, A.C.: No-reference image quality assessment in the spatial domain. IEEE Transactions on image processing  \textbf{21}(12),  4695--4708 (2012)

\bibitem{mou2023t2i}
Mou, C., Wang, X., Xie, L., Zhang, J., Qi, Z., Shan, Y., Qie, X.: T2i-adapter: Learning adapters to dig out more controllable ability for text-to-image diffusion models. arXiv preprint arXiv:2302.08453  (2023)

\bibitem{nichol2022point}
Nichol, A., Jun, H., Dhariwal, P., Mishkin, P., Chen, M.: Point-e: A system for generating 3d point clouds from complex prompts. arXiv preprint arXiv:2212.08751  (2022)

\bibitem{nichol2022glide}
Nichol, A.Q., Dhariwal, P., Ramesh, A., Shyam, P., Mishkin, P., Mcgrew, B., Sutskever, I., Chen, M.: Glide: Towards photorealistic image generation and editing with text-guided diffusion models. In: International Conference on Machine Learning. pp. 16784--16804. PMLR (2022)

\bibitem{oechsle2019texture}
Oechsle, M., Mescheder, L., Niemeyer, M., Strauss, T., Geiger, A.: Texture fields: Learning texture representations in function space. In: Proceedings of the IEEE/CVF International Conference on Computer Vision. pp. 4531--4540 (2019)

\bibitem{po2023compositional}
Po, R., Wetzstein, G.: Compositional 3d scene generation using locally conditioned diffusion. arXiv preprint arXiv:2303.12218  (2023)

\bibitem{podell2023sdxl}
Podell, D., English, Z., Lacey, K., Blattmann, A., Dockhorn, T., M{\"u}ller, J., Penna, J., Rombach, R.: Sdxl: Improving latent diffusion models for high-resolution image synthesis. arXiv preprint arXiv:2307.01952  (2023)

\bibitem{poole2022dreamfusion}
Poole, B., Jain, A., Barron, J.T., Mildenhall, B.: Dreamfusion: Text-to-3{d} using 2d diffusion. ICLR  (2023)

\bibitem{qiu2023richdreamer}
Qiu, L., Chen, G., Gu, X., Zuo, Q., Xu, M., Wu, Y., Yuan, W., Dong, Z., Bo, L., Han, X.: Richdreamer: A generalizable normal-depth diffusion model for detail richness in text-to-3d. arXiv preprint arXiv:2311.16918  (2023)

\bibitem{radford2021learning}
Radford, A., Kim, J.W., Hallacy, C., Ramesh, A., Goh, G., Agarwal, S., Sastry, G., Askell, A., Mishkin, P., Clark, J., et~al.: Learning transferable visual models from natural language supervision. In: International conference on machine learning. pp. 8748--8763. PMLR (2021)

\bibitem{richardson2023texture}
Richardson, E., Metzer, G., Alaluf, Y., Giryes, R., Cohen-Or, D.: Texture: Text-guided texturing of 3d shapes. arXiv preprint arXiv:2302.01721  (2023)

\bibitem{rombach2022high}
Rombach, R., Blattmann, A., Lorenz, D., Esser, P., Ommer, B.: High-resolution image synthesis with latent diffusion models. In: Proceedings of the IEEE/CVF conference on computer vision and pattern recognition. pp. 10684--10695 (2022)

\bibitem{saharia2022photorealistic}
Saharia, C., Chan, W., Saxena, S., Li, L., Whang, J., Denton, E.L., Ghasemipour, K., Gontijo~Lopes, R., Karagol~Ayan, B., Salimans, T., et~al.: Photorealistic text-to-image diffusion models with deep language understanding. Advances in Neural Information Processing Systems  \textbf{35},  36479--36494 (2022)

\bibitem{schuhmann2021laion}
Schuhmann, C., Vencu, R., Beaumont, R., Kaczmarczyk, R., Mullis, C., Katta, A., Coombes, T., Jitsev, J., Komatsuzaki, A.: Laion-400m: Open dataset of clip-filtered 400 million image-text pairs. arXiv preprint arXiv:2111.02114  (2021)

\bibitem{sharma2018conceptual}
Sharma, P., Ding, N., Goodman, S., Soricut, R.: Conceptual captions: A cleaned, hypernymed, image alt-text dataset for automatic image captioning. In: Proceedings of the 56th Annual Meeting of the Association for Computational Linguistics (Volume 1: Long Papers). pp. 2556--2565 (2018)

\bibitem{shi2023mvdream}
Shi, Y., Wang, P., Ye, J., Long, M., Li, K., Yang, X.: Mvdream: Multi-view diffusion for 3d generation. arXiv preprint arXiv:2308.16512  (2023)

\bibitem{siddiqui2023meshgpt}
Siddiqui, Y., Alliegro, A., Artemov, A., Tommasi, T., Sirigatti, D., Rosov, V., Dai, A., Nie{\ss}ner, M.: Meshgpt: Generating triangle meshes with decoder-only transformers. arXiv preprint arXiv:2311.15475  (2023)

\bibitem{siddiqui2022texturify}
Siddiqui, Y., Thies, J., Ma, F., Shan, Q., Nie{\ss}ner, M., Dai, A.: Texturify: Generating textures on 3d shape surfaces. In: European Conference on Computer Vision. pp. 72--88. Springer (2022)

\bibitem{sohl2015deep}
Sohl-Dickstein, J., Weiss, E., Maheswaranathan, N., Ganguli, S.: Deep unsupervised learning using nonequilibrium thermodynamics. In: International conference on machine learning. pp. 2256--2265. PMLR (2015)

\bibitem{song2023roomdreamer}
Song, L., Cao, L., Xu, H., Kang, K., Tang, F., Yuan, J., Zhao, Y.: Roomdreamer: Text-driven 3d indoor scene synthesis with coherent geometry and texture. arXiv preprint arXiv:2305.11337  (2023)

\bibitem{tang2023mvdiffusion}
Tang, S., Zhang, F., Chen, J., Wang, P., Furukawa, Y.: Mvdiffusion: Enabling holistic multi-view image generation with correspondence-aware diffusion. arXiv preprint arXiv:2307.01097  (2023)

\bibitem{voynov2023sketch}
Voynov, A., Aberman, K., Cohen-Or, D.: Sketch-guided text-to-image diffusion models. In: ACM SIGGRAPH 2023 Conference Proceedings. pp. 1--11 (2023)

\bibitem{wang2023score}
Wang, H., Du, X., Li, J., Yeh, R.A., Shakhnarovich, G.: Score jacobian chaining: Lifting pretrained 2d diffusion models for 3d generation. In: Proceedings of the IEEE/CVF Conference on Computer Vision and Pattern Recognition. pp. 12619--12629 (2023)

\bibitem{wang2023breathing}
Wang, T., Kanakis, M., Schindler, K., Van~Gool, L., Obukhov, A.: Breathing new life into 3d assets with generative repainting. arXiv preprint arXiv:2309.08523  (2023)

\bibitem{wang2021real}
Wang, X., Xie, L., Dong, C., Shan, Y.: Real-esrgan: Training real-world blind super-resolution with pure synthetic data. In: Proceedings of the IEEE/CVF international conference on computer vision. pp. 1905--1914 (2021)

\bibitem{wang2023prolificdreamer}
Wang, Z., Lu, C., Wang, Y., Bao, F., Li, C., Su, H., Zhu, J.: Prolificdreamer: High-fidelity and diverse text-to-3d generation with variational score distillation. arXiv preprint arXiv:2305.16213  (2023)

\bibitem{wang2004image}
Wang, Z., Bovik, A.C., Sheikh, H.R., Simoncelli, E.P.: Image quality assessment: from error visibility to structural similarity. IEEE transactions on image processing  \textbf{13}(4),  600--612 (2004)

\bibitem{yang2023dreamspace}
Yang, B., Dong, W., Ma, L., Hu, W., Liu, X., Cui, Z., Ma, Y.: Dreamspace: Dreaming your room space with text-driven panoramic texture propagation. arXiv preprint arXiv:2310.13119  (2023)

\bibitem{yeh2024texturedreamer}
Yeh, Y.Y., Huang, J.B., Kim, C., Xiao, L., Nguyen-Phuoc, T., Khan, N., Zhang, C., Chandraker, M., Marshall, C.S., Dong, Z., et~al.: Texturedreamer: Image-guided texture synthesis through geometry-aware diffusion. arXiv preprint arXiv:2401.09416  (2024)

\bibitem{youwang2023paint}
Youwang, K., Oh, T.H., Pons-Moll, G.: Paint-it: Text-to-texture synthesis via deep convolutional texture map optimization and physically-based rendering. arXiv preprint arXiv:2312.11360  (2023)

\bibitem{yu2023texture}
Yu, X., Dai, P., Li, W., Ma, L., Liu, Z., Qi, X.: Texture generation on 3d meshes with point-uv diffusion. In: Proceedings of the IEEE/CVF International Conference on Computer Vision. pp. 4206--4216 (2023)

\bibitem{zeng2023paint3d}
Zeng, X., Chen, X., Qi, Z., Liu, W., Zhao, Z., Wang, Z., FU, B., Liu, Y., Yu, G.: Paint3d: Paint anything 3d with lighting-less texture diffusion models (2023)

\bibitem{zhang2023text2nerf}
Zhang, J., Li, X., Wan, Z., Wang, C., Liao, J.: Text2nerf: Text-driven 3d scene generation with neural radiance fields. arXiv preprint arXiv:2305.11588  (2023)

\bibitem{zhang2023adding}
Zhang, L., Rao, A., Agrawala, M.: Adding conditional control to text-to-image diffusion models. In: Proceedings of the IEEE/CVF International Conference on Computer Vision. pp. 3836--3847 (2023)

\bibitem{zhang2023scenewiz3d}
Zhang, Q., Wang, C., Siarohin, A., Zhuang, P., Xu, Y., Yang, C., Lin, D., Zhou, B., Tulyakov, S., Lee, H.Y.: Scenewiz3d: Towards text-guided 3d scene composition. arXiv preprint arXiv:2312.08885  (2023)

\end{thebibliography}

\clearpage
% !TEX root = ../main.tex

\appendix

\section{More Implementation Details}
\label{details}
We provide additional implementation details in the following subsections. All of our experiments are conducted on 4 NVIDIA A100 GPUs, and it takes about $90$ minutes to generate a scene.

\subsection{Text Prompt}
Our method takes a room text prompt along with a compositional mesh based on a given room layout as input and aims to synthesize a complete 3D room texture enabling free novel view rendering inside.
Each text prompt is composed of two parts, the style and the description of all the objects in the scene.
During the experiments, we utilize the `Emauromin style' as our default style and also use other styles including `Misc Kawaii', `Anime', `Game Pokemon', `Artstyle Impressionist', and so on, which can be found in this website\footnote{https://stable-diffusion-art.com/sdxl-styles/}.
For better stylization results, we also use the corresponding negative prompt for each style. 
For example, here is one of the text prompts used to generate a bedroom:\\
% \emph{Emauromin style. 720 degrees panorama photo view of a bedroom with oil paintings on the wall. a single-size bed, brown cotton pillows, a wooden bedside table, a wooden wardrobe, an empty bookshelf, a white desk, a chair, a square and flat ceiling lamp hanging on the ceiling. finely detailed, purism, ue5, a computer rendering, minimalism, minimal product design.
% } \\
\emph{\textbf{Prompt}: Emauromin style, a bedroom with oil paintings on the wall, a single-size bed, brown cotton pillows, a wooden bedside table, a wooden wardrobe, an empty bookshelf, a white desk, a chair, a square and flat ceiling lamp hanging on the ceiling. finely detailed, purism, computer rendering, minimalism, minimal product design.}\\
\emph{\textbf{Negative prompt}: blurry, blur, text, watermark, render, 3D, NSFW, nude, CGl, monochrome, B\&W. cartoon, painting, smooth, plasticblurry, low-resolution, deep-fried, oversaturated.}\\
All generated texture of 3D rooms presented in this paper and their corresponding text prompts with one specific style are shown in Fig.~\ref{fig:supp-prompt} and Fig.~\ref{fig:supp-ours-3dfront}. As for the style prompt and the corresponding negative prompt, please refer to the aforementioned website for details.
During the iterative object texturing, the text prompt of every single object is its text description as well as the style instead of the whole text prompt as shown above.

\subsection{Room Geometry Generation}
Though the core of our method lies in generating the texture of a 3D room, it is quite straightforward to combine our method with some 3D shape generators instead of merely utilizing datasets like 3D-FRONT~\cite{fu20213d} designed by professional artists for better convenience and flexibility. Despite the fact that the geometry generated by these methods is not perfect, they can still provide the texturing process with strong geometry priors. Generally, we break down the room geometry generation into two parts: object generation and empty room generation.

\noindent\textbf{3D shape generation.}
As for the furniture items in the scene, we generate most of our object meshes by leveraging an off-the-shelf text-to-3D object generative model, Shap-E\cite{jun2023shap}.
We cut out object descriptions like \emph{`a wooden bedside table'} or \emph{`a white desk'} from the text prompt above, and then send them to the 3D object generative models to generate the corresponding 3D shapes.
Delicate decorations like ceiling lamps and chandeliers are borrowed from Objaverse\cite{deitke2023objaverse} since we found that current object generators are still incapable of generating such fine-grained decorations.
The scarcity of such data in 3D object datasets like ShapeNet\cite{shapenet2015} makes it hard for a 3D generative model to learn.
However, we believe that the quality gap between experts and 3D generators, especially for fine-grained models, will be closed with the rapid development of large-scale 3D generative models.

\noindent\textbf{Empty room generation.}
A procedural generation process is applied to get an empty room mesh.
Based on our observations of indoor scene datasets like 3D-FRONT\cite{fu20213d}, we provide users with various options for diverse room meshes.
Specifically, they can decide whether to include \emph{baseboards}, where to position \emph{doors} and \emph{windows}, which \emph{ceiling style} to choose, and the size of the room.
Under the guidance of these choices,
an empty room mesh can be generated automatically.
For example, the available ceiling styles are illustrated in Fig.~\ref{fig:supp-ceiling}.
Moreover, the generated 3D shapes can also be included in the room according to the provided room layout, and thus a complete room mesh is obtained.

\subsection{Panorama Generation Details}

\noindent\textbf{Initial panorama generation.}
We leverage the SDXL $1.0$ base and refiner models~\cite{podell2023sdxl} for image generation, where the sampler is selected as `Euler a'.
The sampling step is $50$, and we switch from the base model to the refiner at fraction $0.8$, \ie, $40$ steps.
To generate a panoramic image with better visual fidelity and less distortion, we additionally add `720 degrees panorama photo view of' to the beginning of text prompt.
To employ the depth guidance,
a depth-based ControlNet model~\cite{zhang2023adding} is also applied, and the control weight is $1.5$.
Besides, we also use SDXL VAE, and the CFG scale is set to $6.5$.

\begin{figure}[t!]
    \centering
    \includegraphics[width=0.85\linewidth]{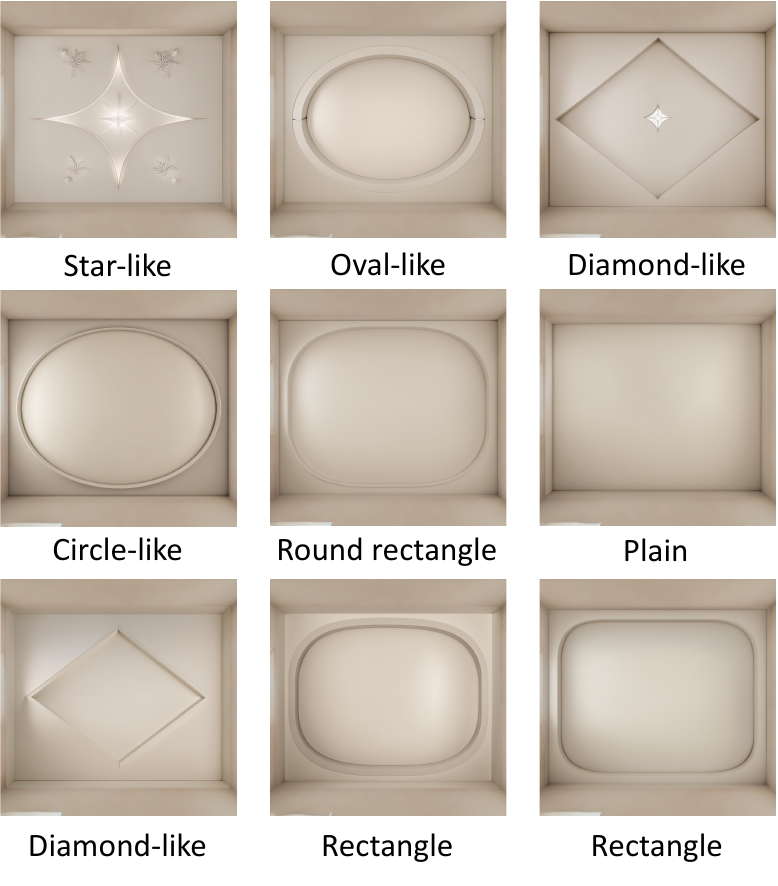}
    \vspace{-8pt}
    \caption{\small \textbf{Different ceiling styles.} We show different designs of ceiling styles with an upward view. }
    \label{fig:supp-ceiling}
\end{figure}

\noindent\textbf{Empty room refinement.}
When refining the ceilings and floors of an empty room,
we will select an upward view and an overhead view to capture the corresponding areas.
The virtual camera is put at the center of the room and towards the center of the ceiling or floor.
% For the upward view, a camera will be placed at the center of the ground towards the center of the room. For the overhead view, another camera will be placed at the center of the ceiling towards the center of the room.
The focal length and the mask will be adjusted according to the width and height of the room. 
As shown in Fig.~\ref{fig:supp-ceiling}, ceilings with star-like or diamond-like decorations and some other styles are all supported in our method.

\begin{figure}[t!]
    \centering
    \includegraphics[width=0.95\linewidth]{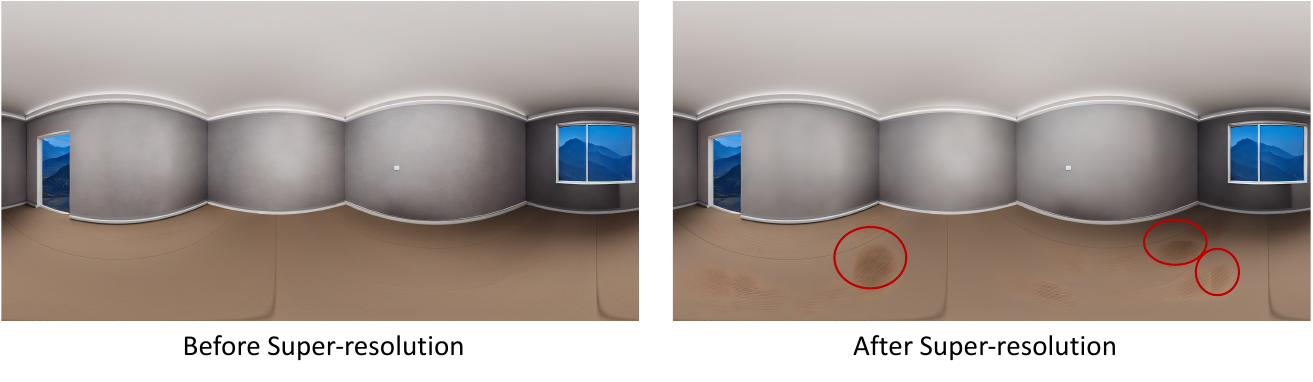}
    %\vspace{-8pt}
    \caption{\small \textbf{Limitation of super-resolution modules.} Some weird artifacts as circled out may appear as shown in the image at the bottom.}
    \label{fig:supp-upscale}
\end{figure}

\noindent\textbf{Super-resolution and its limitation.}
Due to the limitation of memory and inference speed,
the generated image from the SDXL model has a resolution of $2,048 \times 1,024$.
To enrich the texture details, we leverage an off-the-shelf super-resolution method~\cite{wang2021real} to upscale these panorama images to $4,096 \times 2,048$.
Unfortunately, some weird artifacts may appear after using the super-resolution method as illustrated in Fig.~\ref{fig:supp-upscale}. 
% For perspective images, those will first be generated at size $1024\times1024$, and then upscale by 4 times.

% \begin{figure}[t!]
% 	\begin{minipage}{0.47\linewidth}
% 		\centering
%             % \vspace{-3pt}
%             \includegraphics[width=\linewidth]{figs/supp-upscale.pdf}
% 	    \caption{\small \textbf{Limitation of super-resolution modules.} Some weird artifacts as circled out may appear as shown in the image at the bottom.}
%             \label{fig:supp-upscale}
% 	\end{minipage}
% 	\hfill
% 	\begin{minipage}{0.5\linewidth}
% 		\centering
%             %\vspace{-3pt}
% 		\includegraphics[width=\linewidth]{figs/supp_sparse-mask.pdf}
%             %\vspace{-18pt}
% 		\caption{\small \textbf{Ablation study on inpainting methods.} We show comparison results of inpainting using ControlNet only and our method. (a) is the rendering image from a novel view and (b) is the inpainting mask (white area). (c) shows the inpainting result using ControlNet only and (d) shows the inpainting result from our method. We can observe clear messy areas in (c) since the diffusion-based inpainting model is insensitive to sparse masks.}
%             \label{fig:supp-mask}
% 	\end{minipage}
% 	% \vspace{-10pt}
% \end{figure}

\subsection{Iterative Object Texturing Details}
\noindent\textbf{Settings of initial perspective view.}
When re-projecting $\mI_p$ to the initial perspective view $\vv_0$, the default focal length of the virtual camera is set to $500$.
However, if the default setting leads to a bad situation where the object occupies less than half of the image or extends beyond the image boundary, we will adjust the camera's focal length accordingly. To be specific, we would gradually increase the focal length until an object occupies half the width of the image while simultaneously guaranteeing it does not exceed the image.
The resolution of perspective images is $1,024 \times 1,024$, and these images will also be upscaled to $4,096 \times 4,096$ via super-resolution modules.
The setting of SDXL is the same as that for initial panorama generation.

\noindent\textbf{View selection.}
We divide the views used for iterative object texturing into two groups: basic views and additional views.
First, eight basic views are selected and all of them target at the center of the object.
These cameras are roughly located in eight corners of the bounding box covering the whole 3D object.
% The positions of these eight virtual cameras are:
% \begin{equation}\label{eq1}
% \vp_\text{basic}=\left\{(\vx \cdot s + d, \vy \cdot s + d, \vz \cdot s)\right\}
% \end{equation}
% where $\vx, \vy, \vz$ are selected from the boundary value of the object centered at zero. For example, if the object ranges from $-x_0$ to $x_0$ in the $x$ dimension, $\vx$ will be selected from $\left\{-x_0, x_0\right\}$. $s$ is 1.4 and $d$ is 0.15 so that the basic group of views can capture the object. 
For the additional views, different strategies will be applied according to the length-width ratio of the object.
If this ratio is less than $1.5$, eight additional cameras will be used and still target the center of the object.
Their positions are located on a sphere centered around the object with a radius of $0.7$ times the diagonal length of the object bounding box, the elevation angle is set to be a random value between $\pi/6$ and $\pi/3$, and the azimuth angles are set to $0$, $\pi/2$, $\pi$ and $3 \pi/2$, respectively.
Besides, if the aspect ratio is larger than $1.5$, we will select 2 groups of eight additional cameras, \ie, $16$ cameras in total.
In particular, each group of cameras will be also located on a sphere but centered at one-third of the length of this 3D object, ensuring all the objects can be completely viewed with such $16$ cameras.
%
% We may suppose that the object covers a wider range in $x$ dimension, then the two centers will be:
% \begin{equation}\label{eq3}
% \vc = (\pm1/3 \cdot \sqrt{x_0^{2} - y_0^{2}}, 0, 0)
% \end{equation}
% eight virtual cameras will be placed around the sphere for each center.
It's noteworthy that virtual cameras will be strictly placed within the room boundary, and cameras that are too close to the object will be deleted too.

\noindent\textbf{Mask of untextured area.}
As we warp our images to a novel view, those originally occluded parts may be observed due to the sparsity of point clouds.
For example, the front-side texture of a wardrobe may appear when we inpaint its back side.
To eliminate such unreasonable pixels, we identify these areas where the depth is larger than the ground-truth depth and then remove these pixels thereby.
% Moreover, to enable a smoother transition around the border regions, \ie, the area around the boundary of inpainting mask, we will dilate the mask a little bit.

\begin{figure}[t!]
    \centering
    \includegraphics[width=0.95\linewidth]{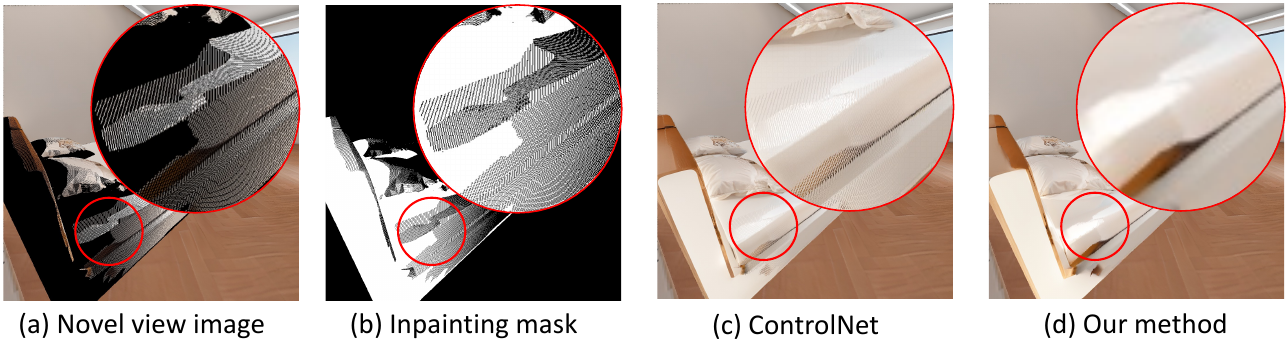}
    %\vspace{-8pt}
    \caption{\small \textbf{Ablation study on inpainting methods.} We show comparison results of inpainting using ControlNet only and our method. (a) is the rendering image from a novel view and (b) is the inpainting mask (white area). (c) shows the inpainting result using ControlNet only and (d) shows the inpainting result from our method. We can observe clear messy areas in (c) since the diffusion-based inpainting model is insensitive to sparse masks.}
    \vspace{0pt}
    \label{fig:supp-mask}
\end{figure}

\noindent\textbf{Inpainting strategy in sparse mask area.}
We use an interpolation-based method to inpaint areas with relatively sparse masks.
Specifically, the interpolation-based method means Telea's inpaint algorithm in OpenCV.
A comparison with using the diffusion model to inpaint these kinds of areas is shown in Fig.~\ref{fig:supp-mask}.
It can be seen that ControlNet does not perform well in sparse areas while our method can generate consistent and natural results.

\noindent\textbf{Selecting satisfying images.}
It is known that images generated by diffusion models exhibit a high degree of diversity, which makes it necessary to select one satisfying image from multiple generation candidates.
While selecting the initial perspective view,
we already have the text prompt $\mT$ and the warped image $\mI_\text{ref}$ from $\mI_p$.
To make sure the generated image aligns with $\mT$ well and is similar to $\mI_\text{ref}$, we compute SSIM Score~\cite{wang2004image} and CLIP Score~\cite{radford2021learning} among the candidate images and select the one with the highest score:
\begin{equation}\label{eq4}
\mI_\text{obj} = \mathop{\arg\max}_{j}(\mS(\mI_\text{obj}^{j}, \mI_\text{ref}) + \mC(\mI_\text{obj}^{j}, \mT))
\end{equation}
where $\mS(\cdot)$ is the function to calculate SSIM Score, $\mC(\cdot)$ stands for the function to calculate CLIP Score and $\left\{\mI_\text{obj}^{j}\right\}_{j=0}^{5}$ represent 5 candidate images used here.

During the iterative object texturing,
we notice that the inpainted areas can't strictly align with other regions on the perspective images, leading to inconsistent styles and weird patterns.
Hence, we will dilate the inpainting mask and leverage the dilation areas to judge the style consistency.
% apart from the image quality, we should also consider the 3D consistency between different views.
% We notice that the unmasked area of the original image won't be exactly the same as that after inpainting, even under the guidance of the inpainting mask, which significantly damages the 3D consistency especially in the dilated mask area as mentioned before.
Specifically, we additionally compute a PSNR score in the dilated area to encourage good alignment following:
\begin{equation}\label{eq5}
\mI_\text{obj, i} = \mathop{\arg\max}_{j}(\mP(\mI_\text{obj, i}^{j}, \mI_\text{ref, i}) + \mC(\mI_\text{obj, i}^{j}, \mT))
\end{equation}
where $\mP(\cdot)$ represents the function to calculate the PSNR score, $\mI_\text{ref, i}$ is the image to be inpainted under view $\vv_i$, and the CLIP score is also used to select the most suitable inpainting results from 5 candidates.

\subsection{Fine-grained Texture Control}
\begin{figure*}[t!]
    \centering
    \includegraphics[width=0.9\linewidth]{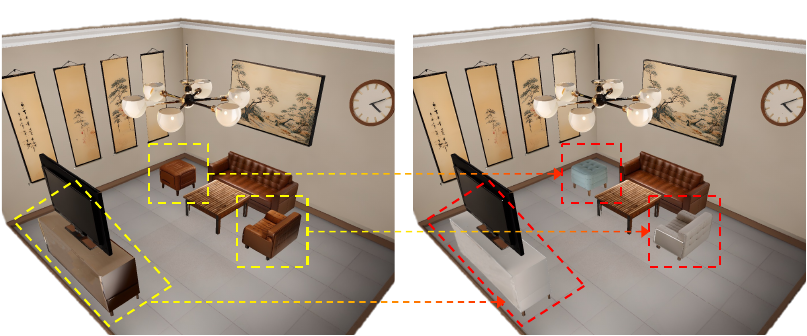}
    \caption{\small \textbf{Object-level alignment results.} We show an object-level alignment result in the living room by simply ignoring the reference texture from the initial panorama. Every object in this scene aligns with its corresponding text prompt.}
    \label{fig:aligning}
\end{figure*}

\begin{figure*}[t!]
    \centering
    \includegraphics[width=0.9\linewidth]{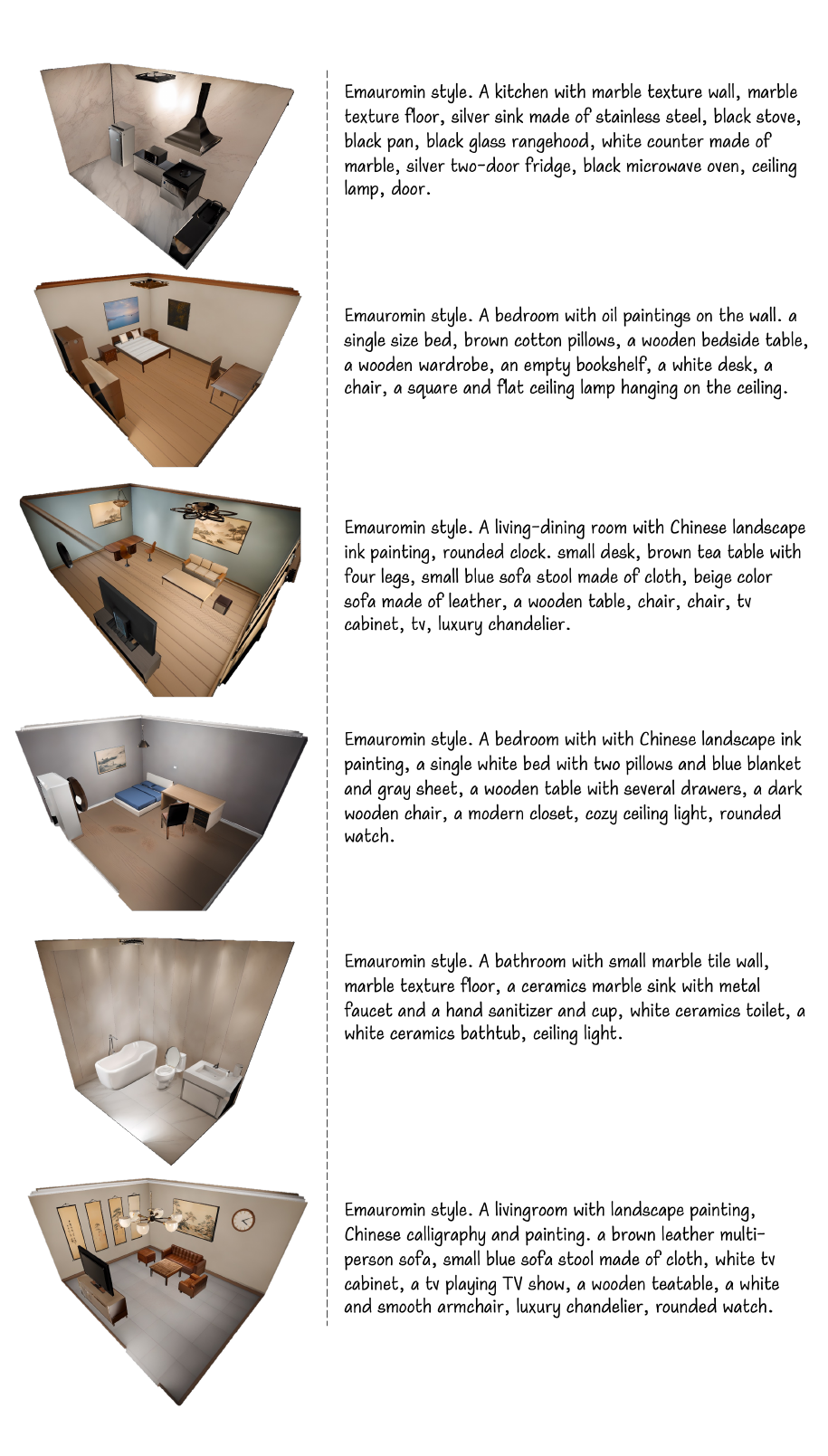}
    \vspace{-16pt}
    \caption{\small \textbf{Generated rooms and their corresponding text prompts.} 6 compositional rooms with default style are shown with an overview image on the left and their corresponding text prompts on the right.}
    \label{fig:supp-prompt}
\end{figure*}

Since our method aims to generate harmonious texture across the whole scene, it is natural for our method to ignore some semantics in the object-level text if they break the overall consistency significantly(like a blue stool among a bunch of brown furniture in the last example of Fig.~\ref{fig:supp-prompt}). This misalignment is mainly due to the SDXL model, which is trained on real-world scene images with globally consistent textures. However, it is easy for our method to align with all the object-level prompts by sacrificing some extent of harmoniousness. It is up to the users themselves to decide how they would like the room texture. This fine-grained control can be achieved by simply ignoring the reference textures of these objects in the panorama during the object texturing process. Such a compromised result is shown in the teaser image as well as in Fig.~\ref{fig:aligning}. Apart from aligning object textures perfectly with text prompts, other fine-grained texture controls including controlling the texture of floors, walls, ceilings, and objects using scribbles are also integrated into one scene in the demo video.

\section{User Study}

\begin{figure*}[t!]
    \centering
    \includegraphics[width=0.9\linewidth]{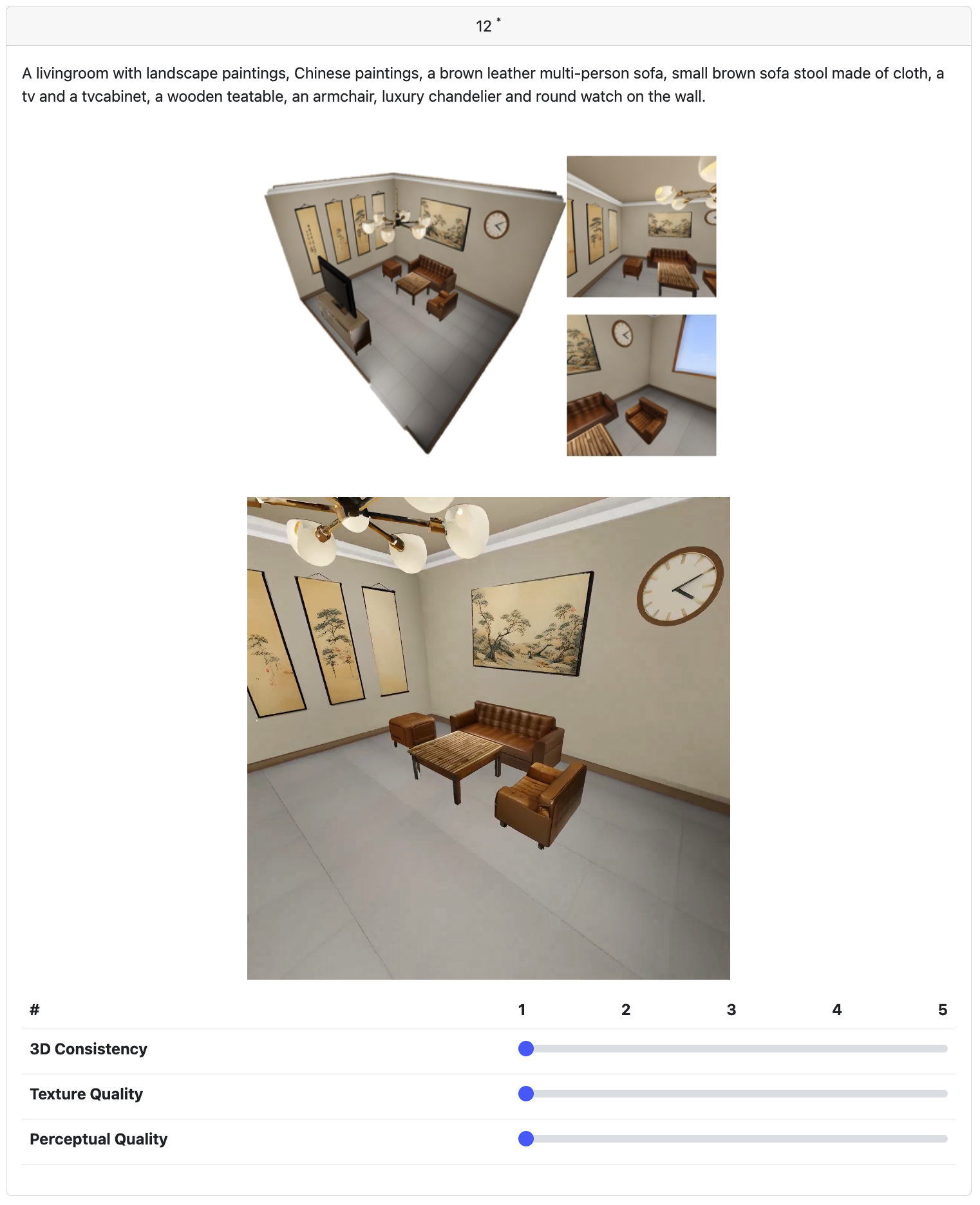}
    \caption{\small \textbf{The interface of our questionnaire used in the user study.} The text prompt is shown on the top, the room overview and two randomly selected perspective images are in the middle, and a room tour video is put at the bottom.}
    \label{fig:supp-user}
\end{figure*}

We leverage a flask-based web application for the user study to compare our method with baselines from the human perspective.
Fig.~\ref{fig:supp-user} shows the interface of our questionnaire, where the text description is put on the top, the room overview(a top view of the room) and two random perspective images are in the middle, and a video showing free roaming in the room is also provided below.
In the questionnaires, we have 6 groups of scenes in total, where 3 results from baselines and 1 from ours are included in each group.
We invite $61$ volunteers to conduct the user study and each participant will be randomly shown 2 groups of scenes, \ie, 8 generated scenes, and be asked to judge each presented scene from three different dimensions, 3D consistency(3DC), texture quality(TQ), and perceptual quality(PQ).
Specifically, they have to give a score ranging from 1 to 5 for such three aspects. The higher, the better.
In the end, we gather $488$ responses from the $61$ participants and calculate the overall preferences as shown in Tab.~\ref{tab:supp_user}. 

\begin{table}[!t]
\begin{center}
\footnotesize

\begin{tabular}{lccccc}
\hline
Method &3DC($\uparrow$) &TQ($\uparrow$) &PQ($\uparrow$) \\
\hline
TEXTure-C~\cite{richardson2023texture} & 3.06($\pm$0.85)  & 2.75($\pm$0.80)  & 2.88($\pm$0.77) \\ 
TEXTure-H~\cite{richardson2023texture} & 2.83($\pm$0.86)  & 2.63($\pm$0.84)  & 2.64($\pm$0.83) \\
SceneTex~\cite{chen2023scenetex}  & 3.98($\pm$0.86)  & 3.73($\pm$1.02)  & 3.56($\pm$0.95) \\ 
Ours & \textbf{4.51}($\pm$0.71) & \textbf{4.29}($\pm$0.81) & \textbf{4.26}($\pm$0.76) \\ 
\hline
\end{tabular}

\end{center}
% \vspace{-12pt}
\caption{
\textbf{Quantitative comparison of the user study.} Mean opinion scores are in the range of 1 $\sim$ 5. Our method outperforms TEXTure-C and TEXTure-H by a large margin.
}

\label{tab:supp_user}
\end{table}

\begin{figure*}[t!]
    \centering
    \includegraphics[width=0.95\linewidth]{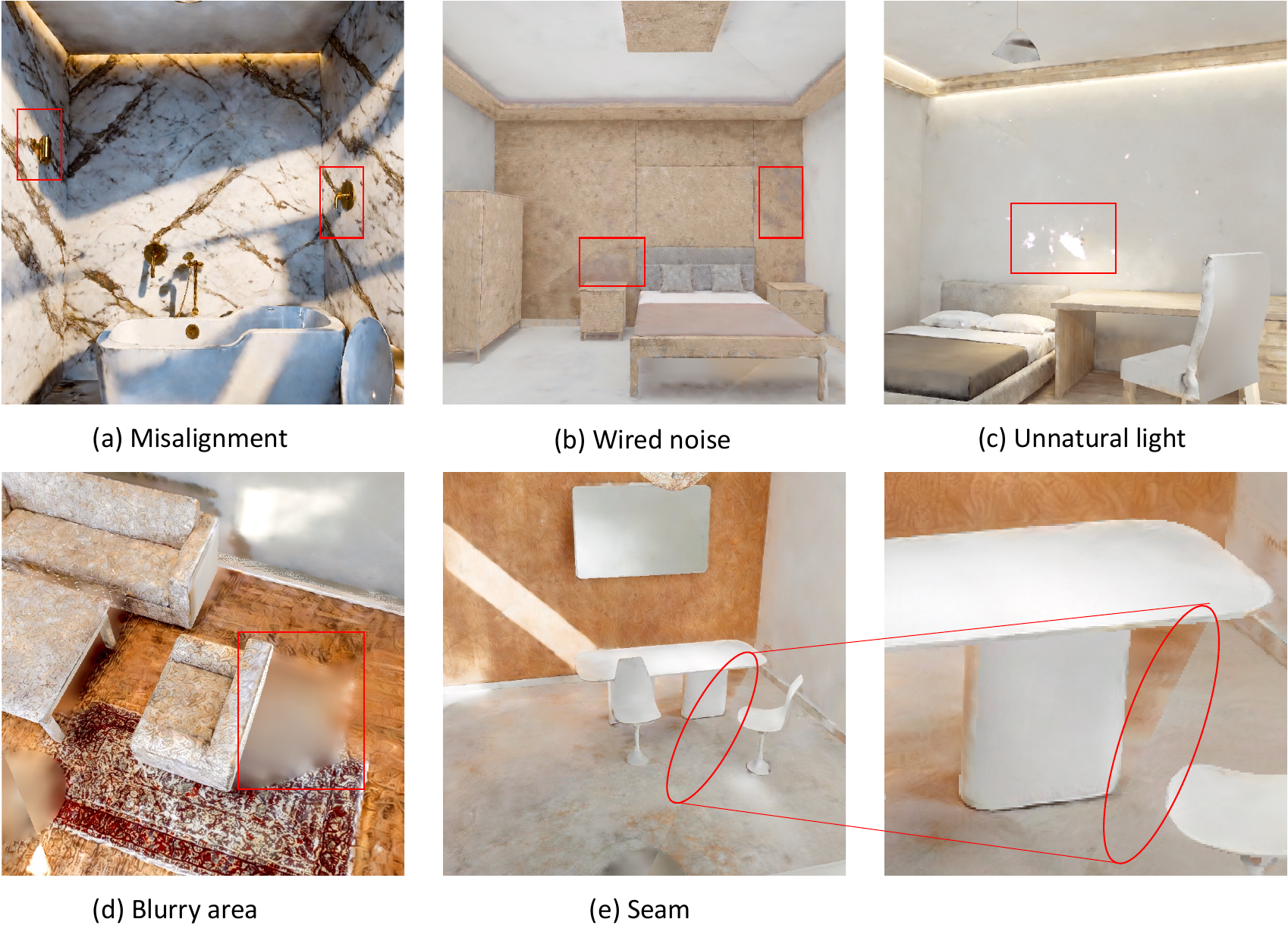}
    % \vspace{-12pt}
    \caption{\small \textbf{Limitations of optimization-based framework.} We show several obvious artifacts of the optimization-based framework including misalignment between geometry and texture, the existence of unnatural light and noise, and evident blurry areas and seams. }
    \label{fig:supp-scenetexl}
\end{figure*}

\section{Detailed Comparison with SceneTex}
As the most closely related and concurrent work as ours, SceneTex~\cite{chen2023scenetex} formulates the whole texturing process as an optimization problem by using a multi-resolution texture field and the VSD objective~\cite{wang2023prolificdreamer}. Though being able to generate compelling textures for a given room geometry, the optimization-based framework may suffer some underlying problems. First of all, the texture generated via optimization may not align well with the given geometry as the texture prior distilled from text-to-image diffusion models tends to make images look as realistic as possible from certain viewpoints.
For example, as shown in Fig.~\ref{fig:supp-scenetexl} (a), there should not exist handle-like objects on the walls of the bathroom since there are no handles at all in the given meshes.
Similarly, SceneTex is prone to generate indoor textures containing unnatural lights and weird noises as shown in Fig.~\ref{fig:supp-scenetexl} (b) and (c).
Moreover, the choice of viewpoints leads to some blurry areas due to the severe occlusion problem in the indoor scene as shown in Fig.~\ref{fig:supp-scenetexl} (d). However, we believe the blurry problem may be mitigated via a more delicate viewpoint selection strategy.
Some clear seams can be observed in Fig.~\ref{fig:supp-scenetexl} (e) due to the usage of UV map. 
%noise, light, wired pattern, blurry due to occlusion
%those need extremely fancy style may be more fit
%应该要说一下他们方法更适用的范围？
% On the other hand, for objects with complex topological structures like multi-layer lamps, our inpainting-based method may not be a perfect solution due to the lack of viewpoints observing the inner surfaces, while an optimization-based framework may be a better choice as it provides a smoother texture space.
On the other hand, textures of objects with complex topological structures may easily be affected by accumulative errors under an explicit inpainting-based framework, even though we have designed a module to detect the misalignment between depth space and rgb space. Optimization-based methods naturally possess some extent of continuity and will not be significantly impacted by a particular viewpoint. But objects with complex topological structures like multi-layer lamps still pose a challenge for both approaches due to the severe self-occlusion.
In the future, we believe a well-designed strategy could marry the merits of the inpainting-based method and the optimization-based method for more harmonious and consistent texture generation.

\section{Additional Results}

\begin{figure*}[t!]
    \centering
    \includegraphics[width=0.9\linewidth]{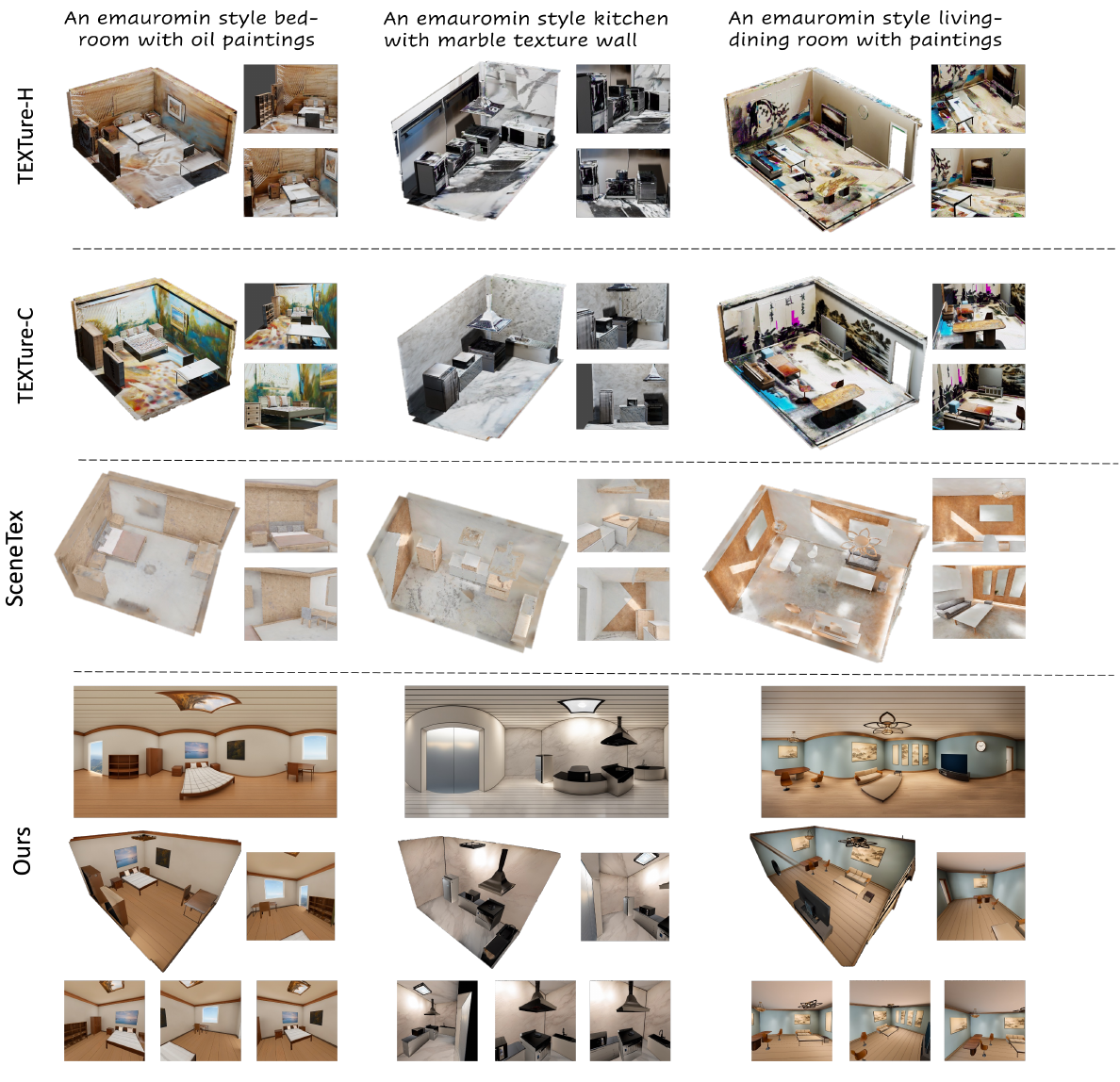}
    \caption{\small \textbf{More qualitative comparison.} We show more qualitative results compared with baselines.}
    \vspace{-12pt}
    \label{fig:supp-quali}
\end{figure*}

\begin{figure*}[t!]
    \centering
    \includegraphics[width=0.9\linewidth]{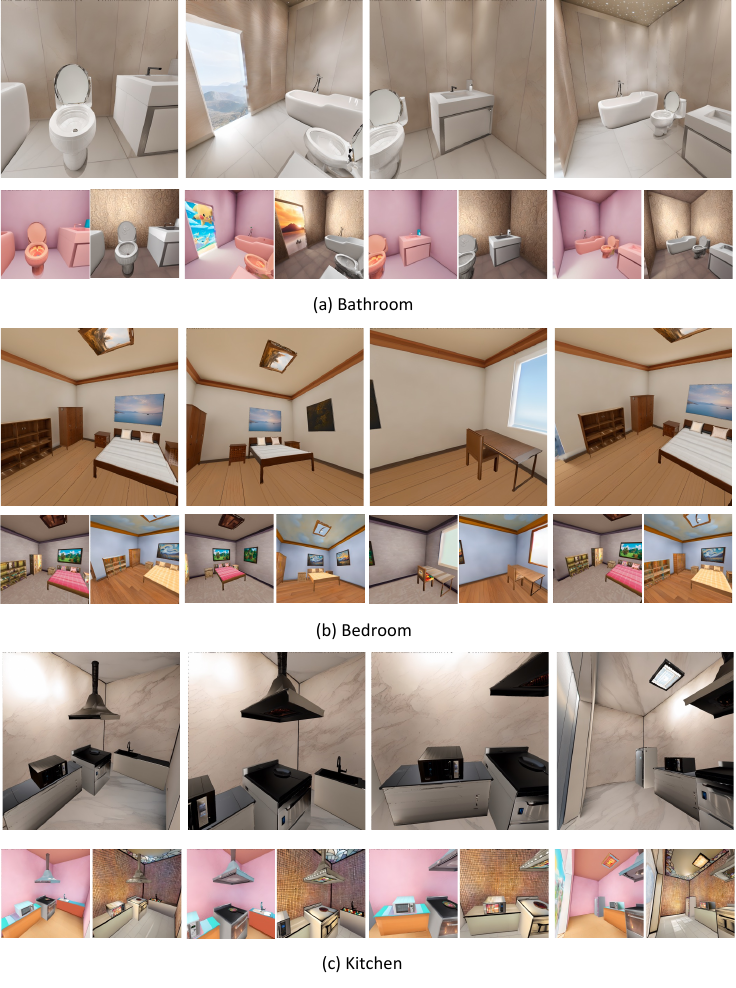}
    \vspace{-12pt}
    \caption{\small \textbf{Results of stylized rooms.} We show some stylized rooms with several rendered perspective images from several perspective views.}
    \label{fig:supp-style}
\end{figure*}

\begin{figure*}[t!]
    \centering
    \includegraphics[width=0.9\linewidth]{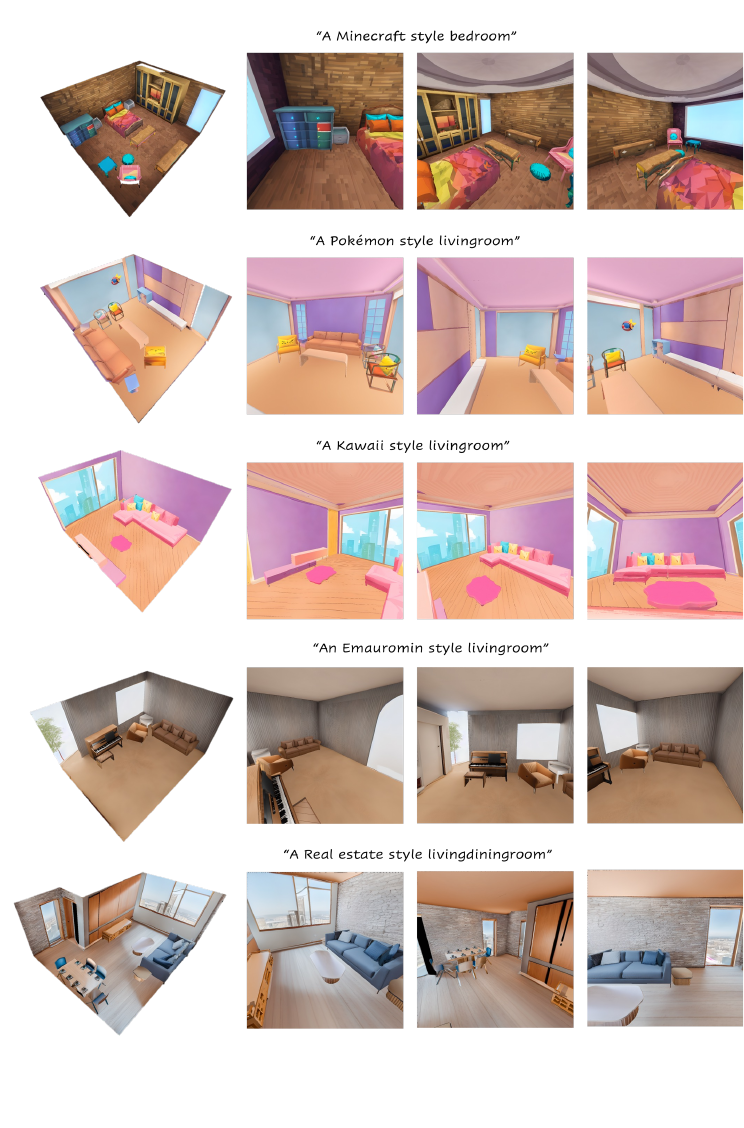}
    \vspace{-12pt}
    \caption{\small \textbf{Our scene texturing results on 3D-FRONT dataset} We show our texturing results on the 3D-FRONT dataset with an overhead view on the left and three perspective views from inside on the right. The text prompt here is concise, please refer to Fig.~\ref{fig:supp-ours-3dfront} for details.}
    \label{fig:supp-ours-3dres}
\end{figure*}

\begin{figure*}[t!]
    \centering
    \includegraphics[width=0.85\linewidth]{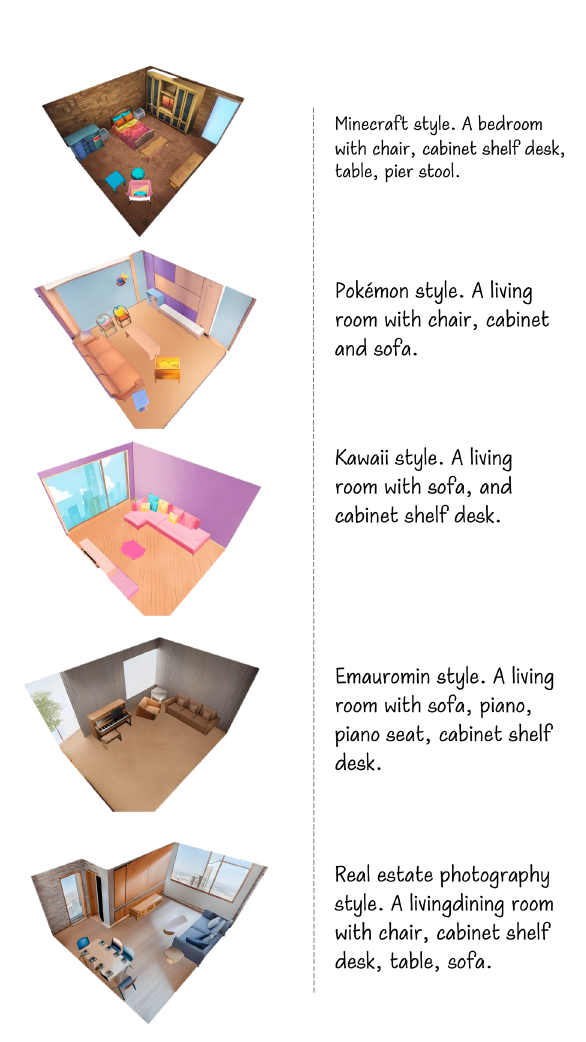}
    \vspace{-12pt}
    \caption{\small \textbf{Our scene texturing results and corresponding text prompts on the 3D-FRONT dataset.} 5 3D-FRONT rooms with different styles are shown with an overview image on the left, and their corresponding text prompts on the right.
}
    \label{fig:supp-ours-3dfront}
\end{figure*}

More qualitative results including a kitchen, a bedroom, and a living-dining room compared with baseline methods are shown in Fig.~\ref{fig:supp-quali} and more stylized room results are shown in Fig.~\ref{fig:supp-style}. Though it is more flexible to assemble a room using 3D shape generators along with our provided empty room generator by users themselves, our method is also capable of texturing a room from professional datasets like 3D-FRONT~\cite{fu20213d}. We choose five rooms including a bedroom, three living rooms, and a living-dining room from the dataset, and the results of the overhead view and several perspective views from inside are shown in Fig.~\ref{fig:supp-ours-3dres}. The overview images of these rooms as well as their corresponding text prompts are shown in Fig.~\ref{fig:supp-ours-3dfront}.
We render some room tour videos of different scenes with different styles, which are integrated into a unified video put in the supplementary. 
Besides, we also present a demo video to demonstrate the effectiveness of using our misalignment detection technique. Another demo video shows how our method supports interactive fine-grained texture controls as well as a room tour video in the new room after applying these controls.

% ---- Bibliography ----
%
% BibTeX users should specify bibliography style 'splncs04'.
% References will then be sorted and formatted in the correct style.
%
% \bibliographystyle{splncs04}
% \bibliography{main}
% \end{document}

\end{document}